\documentclass[nohyperref]{article}

\usepackage{microtype}
\usepackage{graphicx}
\usepackage{caption}
\usepackage[export]{adjustbox}
\usepackage{subcaption}
\usepackage{booktabs} 
\usepackage{svg} 
\usepackage{bm}
\usepackage{multirow}

\usepackage{hyperref}
\usepackage{color}

\usepackage[accepted]{icml2022}

\usepackage{amsmath}
\usepackage{amssymb}
\usepackage{mathtools}
\usepackage{amsthm}

\usepackage[capitalize,noabbrev]{cleveref}

\newcommand{\rulesep}{\unskip\ \vrule\ }

\theoremstyle{plain}
\newtheorem{theorem}{Theorem}[section]

\theoremstyle{definition}
\newtheorem{definition}[theorem]{Definition}

\newtheorem{conjecture}[theorem]{Conjecture}
\theoremstyle{remark}

\usepackage[textsize=tiny]{todonotes}

\icmltitlerunning{Exploring and Exploiting Hubness Priors for High-Quality GAN Latent Sampling}

\begin{document}

\twocolumn[
\icmltitle{Exploring and Exploiting Hubness Priors for High-Quality GAN \\ Latent Sampling}




\begin{icmlauthorlist}
\icmlauthor{Yuanbang Liang}{yyy}
\icmlauthor{Jing Wu}{yyy}
\icmlauthor{Yu-Kun Lai}{yyy}
\icmlauthor{Yipeng Qin}{yyy}
\end{icmlauthorlist}

\icmlaffiliation{yyy}{School of Computer Science and Informatics, Cardiff University, Cardiff, CF24 4AG, UK}

\icmlcorrespondingauthor{Yipeng Qin}{qiny16@cardiff.ac.uk}

\icmlkeywords{Machine Learning, Computing Vision, GAN}

\vskip 0.3in
]



\printAffiliationsAndNotice{}  

\begin{abstract}
Despite the extensive studies on Generative Adversarial Networks (GANs), how to reliably sample high-quality images from their latent spaces remains an under-explored topic.
In this paper, we propose a novel GAN latent sampling method by exploring and exploiting the {\it hubness priors} of GAN latent distributions. 
Our key insight is that the high dimensionality of the GAN latent space will inevitably lead to the emergence of {\it hub} latents that usually have much larger sampling densities than other latents in the latent space.
As a result, these {\it hub} latents are better trained and thus contribute more to the synthesis of high-quality images.
Unlike the a posteriori ``cherry-picking'', our method is highly efficient as it is an a priori method that identifies high-quality latents before the synthesis of images.
Furthermore, we show that the well-known but purely empirical truncation trick is a naive approximation of the central clustering effect of {\it hub} latents, which not only uncovers the rationale of the truncation trick, but also indicates the superiority and fundamentality of our method.
Extensive experimental results demonstrate the effectiveness of the proposed method.
Our code is available at: \url{https://github.com/Byronliang8/HubnessGANSampling}.
\end{abstract}

\section{Introduction}

\begin{figure}[t]
  \centering
    \includegraphics[width=.9\linewidth]{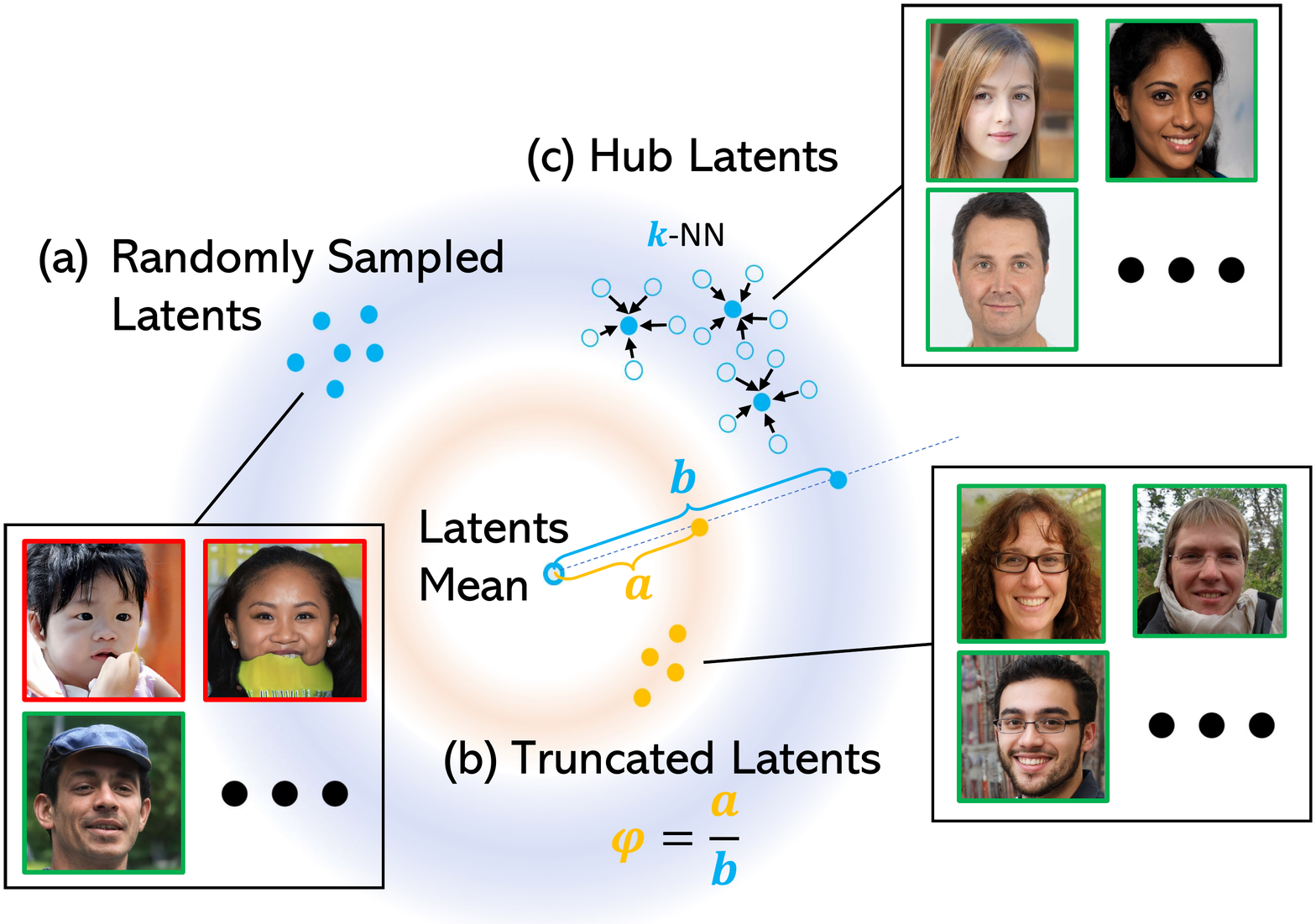}
    \caption{Our method {\it vs.} random latent sampling and the truncation trick~\cite{marchesi2017megapixel,brock2018large,karras2019style}. All images are generated using StyleGAN2~\cite{karras2020analyzing}. (a) Random latent sampling yields both high-quality (\textcolor{green}{green} box) and low-quality (\textcolor{red}{red} box) images; (b) The truncation trick improves the quality of synthesized images by empirically truncating randomly sampled latents according to a scaling parameter $\psi$ ({\it e.g.} $\psi=0.7$), which is a naive approximation of the ``central clustering effect'' of our {\it hub} latents; (c) Our method identifies high-quality latents as the {\it hub} latents that are more likely to be among the $k$-nearest neighbors of other latents~\cite{radovanovic2010hubs}. The \textcolor{blue}{blue} and \textcolor{orange}{orange} rings illustrate the high-dimensional Gaussian (latent) distribution~\cite{menon2020pulse} and their truncated version respectively.}
    \label{fig:teaser}
\end{figure}

Generative adversarial networks (GANs) are a type of deep generative models that have revolutionized a variety of applications in computer vision and computer graphics, {\it e.g.} image synthesis~\cite{karras2019style,park2019semantic,zhu2020sean}, image editing~\cite{abdal2019image2stylegan,abdal2020image2stylegan++,tov2021designing}, image-to-image translation~\cite{isola2017image,zhu2017unpaired,richardson2021encoding}.
Among them, novel image synthesis via random latent sampling is the most fundamental. 
It not only generates novel instances from the data distribution, but also measures how close the learned distribution is to the data distribution.
Through the lens of the quality of synthesized images, we have witnessed significant progress in GANs over the past several years.
Specifically, starting from the groundbreaking vanilla GAN \cite{goodfellow2014generative}, DCGAN~\cite{radford2015unsupervised} laid the foundation for GAN architectures as deep convolutional neural networks; ProGAN~\cite{karras2018progressive} showed that GANs can generate high-quality images at high resolutions; BigGAN~\cite{brock2018large} addressed the problem of class-conditional image synthesis; the StyleGAN series~\cite{karras2019style,karras2020analyzing,karras2021alias} further boosted the quality and controllability of synthesized images with their style-based generator architectures and several novel techniques.

Nevertheless, with such improvements, the quality variance among images generated by randomly sampled latents has become increasingly striking (Fig.~\ref{fig:teaser}).
Without curation, the quality of GAN synthesized images can occasionally be very low, which hinders the deployment of GANs in real-world applications.
As a naive solution, ``cherry-picking'' is commonly used to select high-quality images from those synthesized with randomly sampled latents in an {\it a posteriori} manner.
However, in the absence of reliable quantitative measures of the quality of a single GAN-synthesized image\footnote{Existing quantitative measures like FID and Inception scores are all statistical ones that are only applicable to distributions.}, existing ``cherry-picking'' methods are barely manual, thereby being tedious and unscalable.
Addressing this issue, the well-known ``truncation trick''~\cite{marchesi2017megapixel,brock2018large,karras2019style} was proposed, which ``truncates'' randomly sampled GAN latents towards their mean based on the observation that the images synthesized from close-to-mean latents are usually of higher quality.
Although effective, the truncation trick is a purely empirical ``trick'' that brings few new insights to the community.

In this paper, we propose a novel latent sampling method for GANs by exploring and exploiting the {\it hubness} phenomenon~\cite{radovanovic2010hubs} in their latent spaces, which facilitates their synthesis of high-quality images in an {\it a priori} manner.
Specifically, our key insights include: i) the high dimensionality of the GAN latent space will inevitably lead to the emergence of {\it hub} latents that are much more likely to be among the nearest neighbors of other latents in the latent space, {\it i.e.} the hubness phenomenon; 
ii) in general, the quality of a GAN synthesized image is positively correlated with the {\it hub value} of its corresponding latent, {\it i.e.} the number of times a latent becomes a $k$-nearest neighbor ($k$-NN) of other latents in a given latent sample set.
We believe that this positive correlation originates from the well-known close relationship between $k$-NN and density estimation. 
In other words, a higher {\it hub value} usually indicates a higher sampling density, which has a positive effect on the training and thus the quality of synthesized images.
Therefore, we formulate the above insights as the proposed {\it hubness priors} and propose a corresponding method to sample high-quality GAN latents that yield high-quality synthesized images.
Compared to ``cherry-picking'', our method is highly efficient as it is {\it a priori} ({\it i.e.} our high-quality latents are determined before the synthesis of images) and automatic ({\it i.e.} with little human-intervention).
Furthermore, we show that the well-known truncation trick is a naive approximation of the ``central clustering effect''  of our {\it hub} latents~\cite{radovanovic2010hubs}. 
This not only uncovers the rationale of the truncation trick, but also indicates that our method is superior and more fundamental.
Extensive experimental results demonstrate the effectiveness of the proposed method.

In summary, our contributions include:
\begin{itemize}
    \item We uncover the existence of {\it hubness} phenomenon in the GAN latent space, which has a significant correlation with the quality of GAN synthesized images, {\it i.e.} the proposed {\it hubness priors}.
    \item We propose a novel GAN latent sampling algorithm that identifies high-quality {\it hub} latents based on our {\it hubness priors}, which allows efficient and high-quality image synthesis for GANs.
    \item We show that the well-known truncation trick is a naive approximation of the ``central clustering effect'' of our {\it hub} latents. This not only uncovers the rationale of the truncation trick, but also indicates that our method is superior and more fundamental.
\end{itemize}

\begin{figure*}[htp]
 \centering
 \begin{subfigure}[h]{\textwidth}
    \centering
    \includegraphics[width=.32\linewidth]{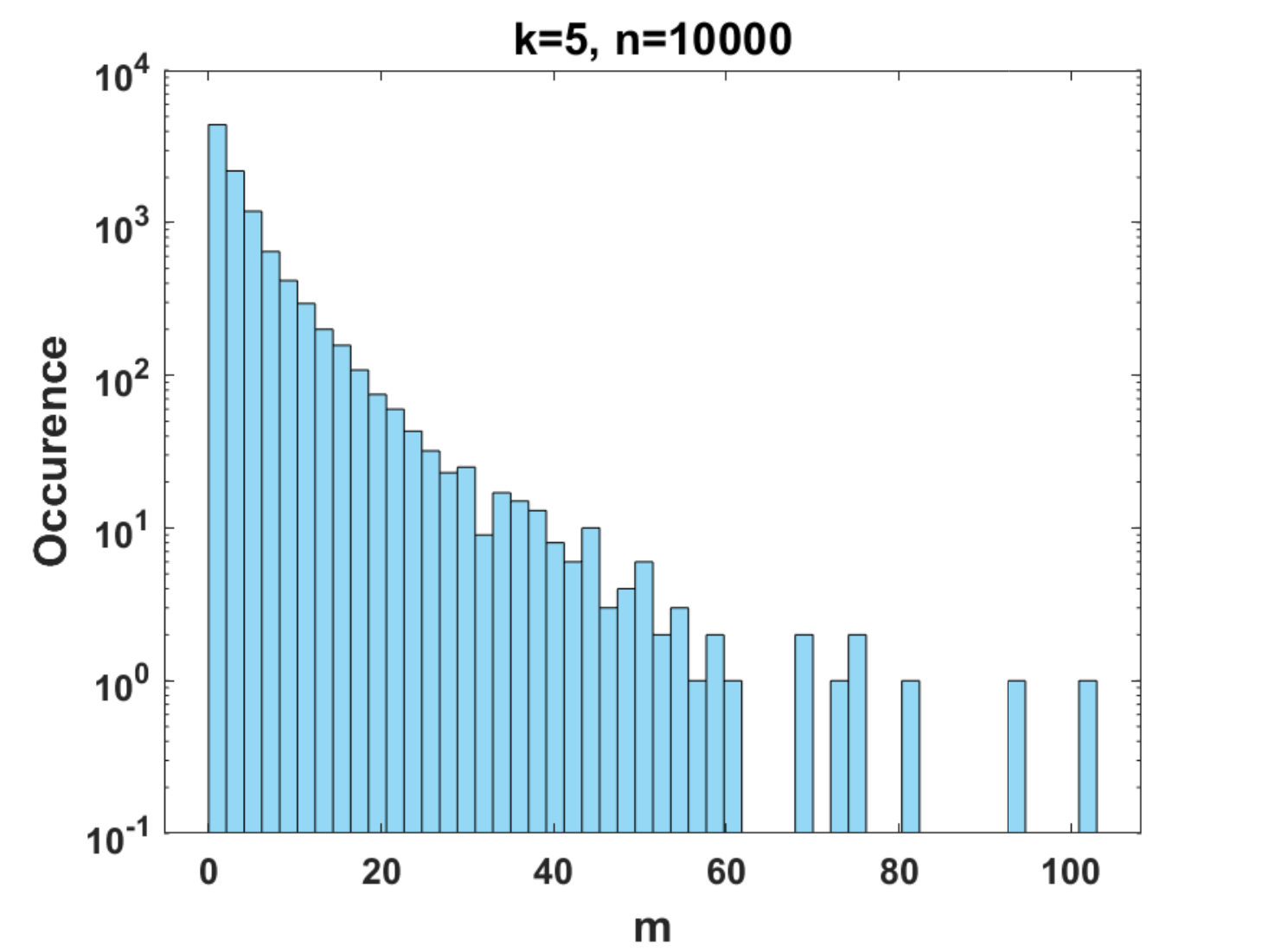}%
    \includegraphics[width=.32\linewidth]{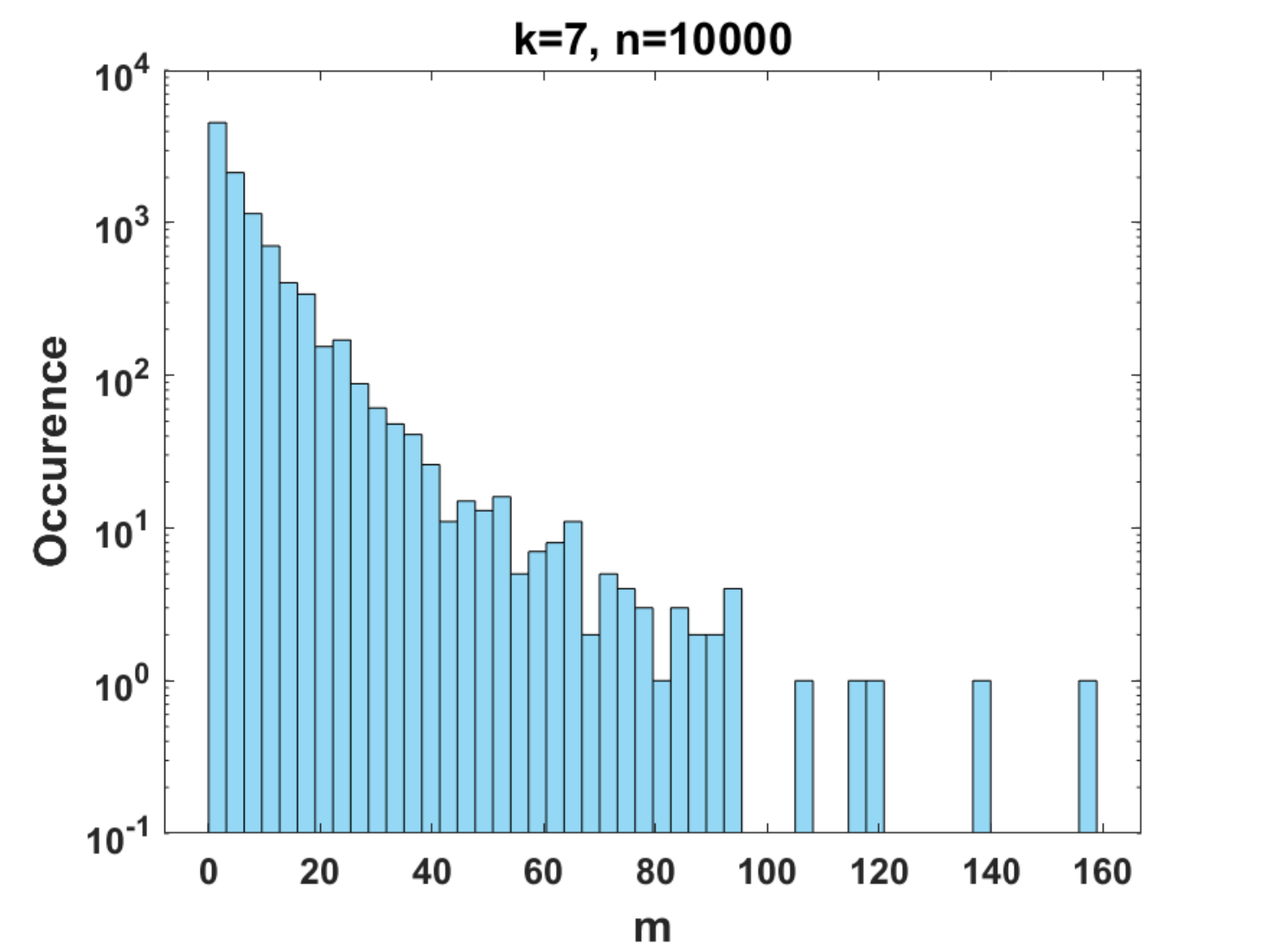}%
   \includegraphics[width=.32\linewidth]{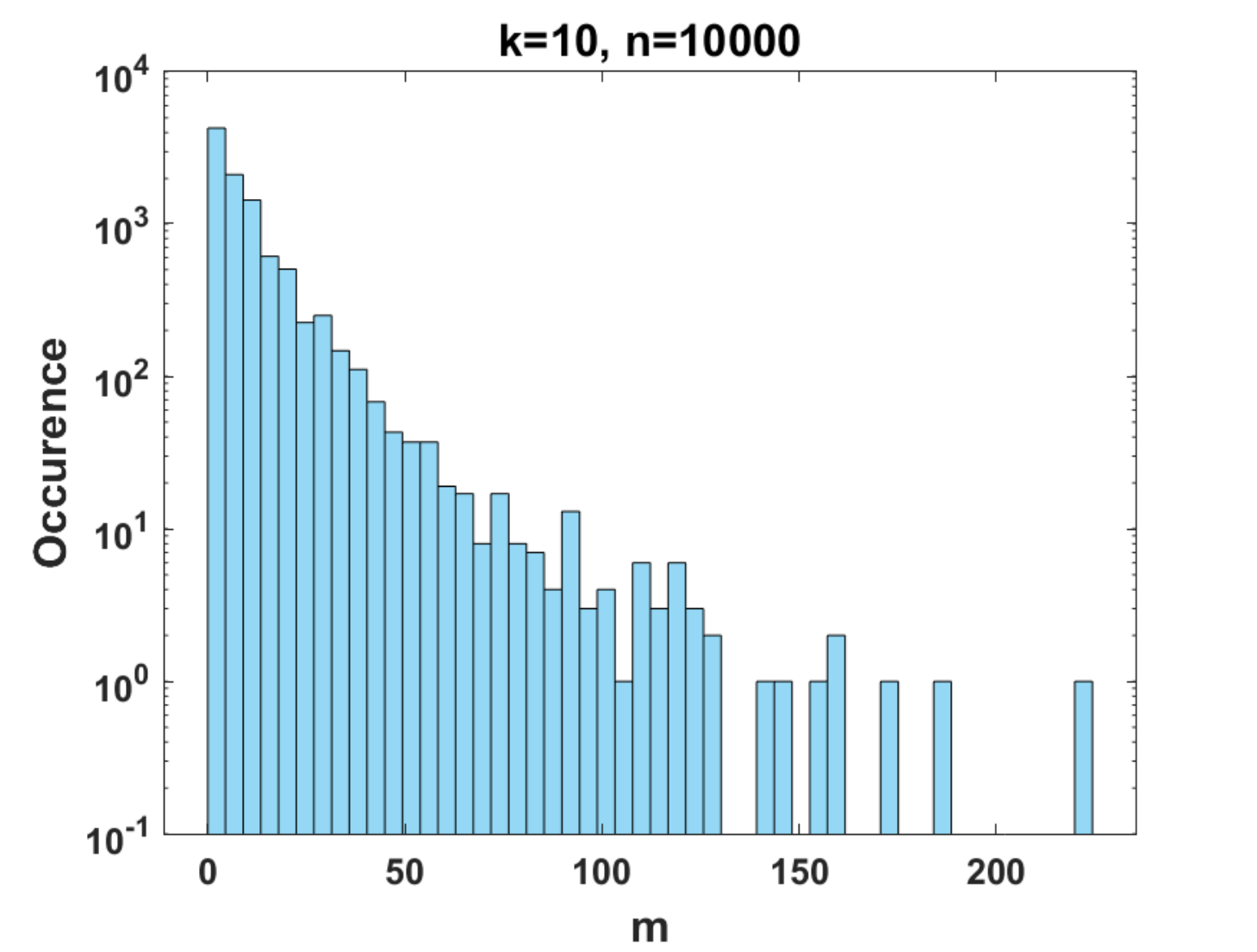}%
   \caption{StyleGAN series~\cite{karras2019style,karras2020analyzing,karras2021alias}, $W$-space ($512$ dimensions)}
 \end{subfigure}
 \begin{subfigure}[h]{\textwidth}
    \centering
    \includegraphics[width=.32\linewidth]{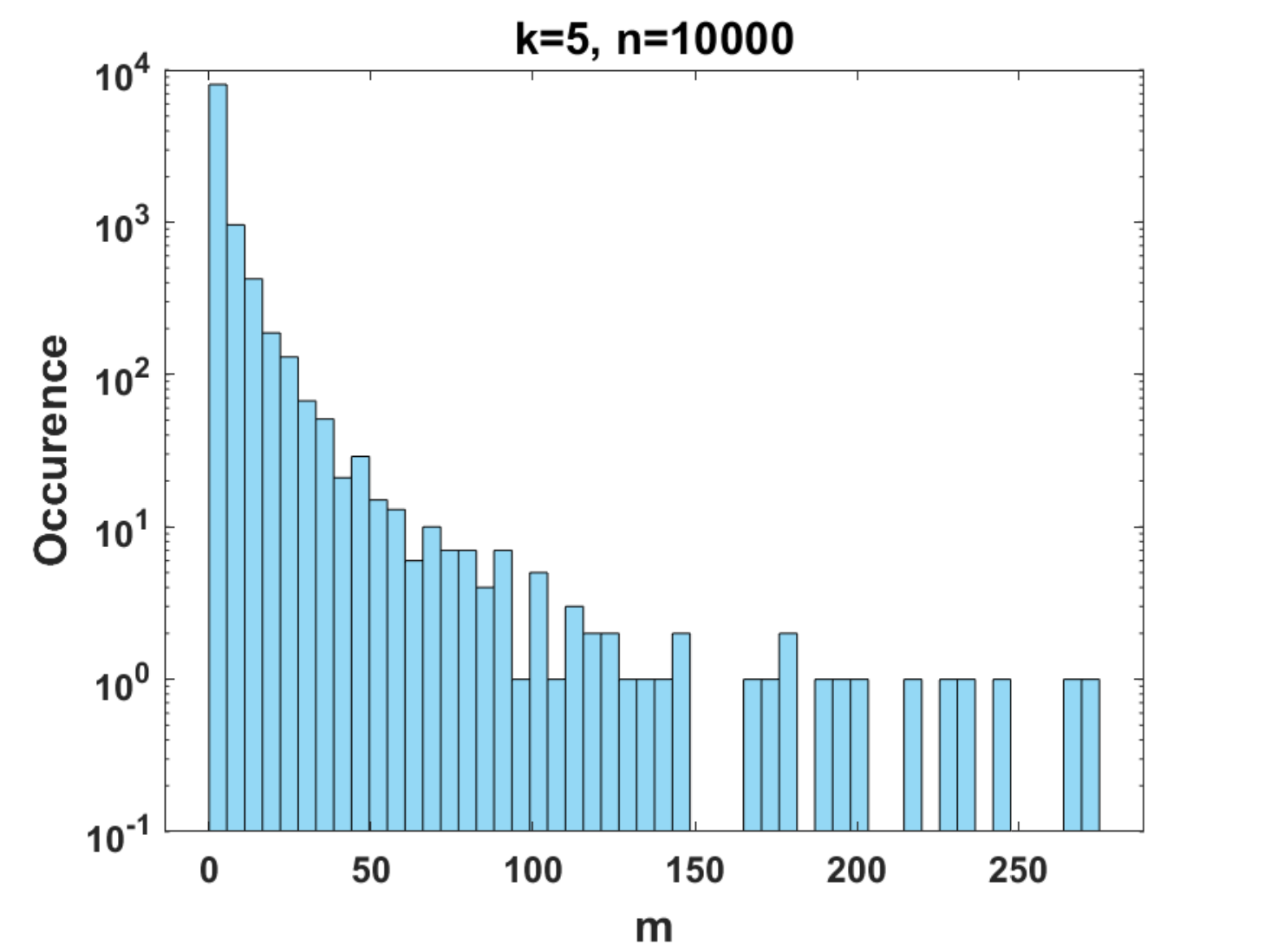}%
    \includegraphics[width=.32\linewidth]{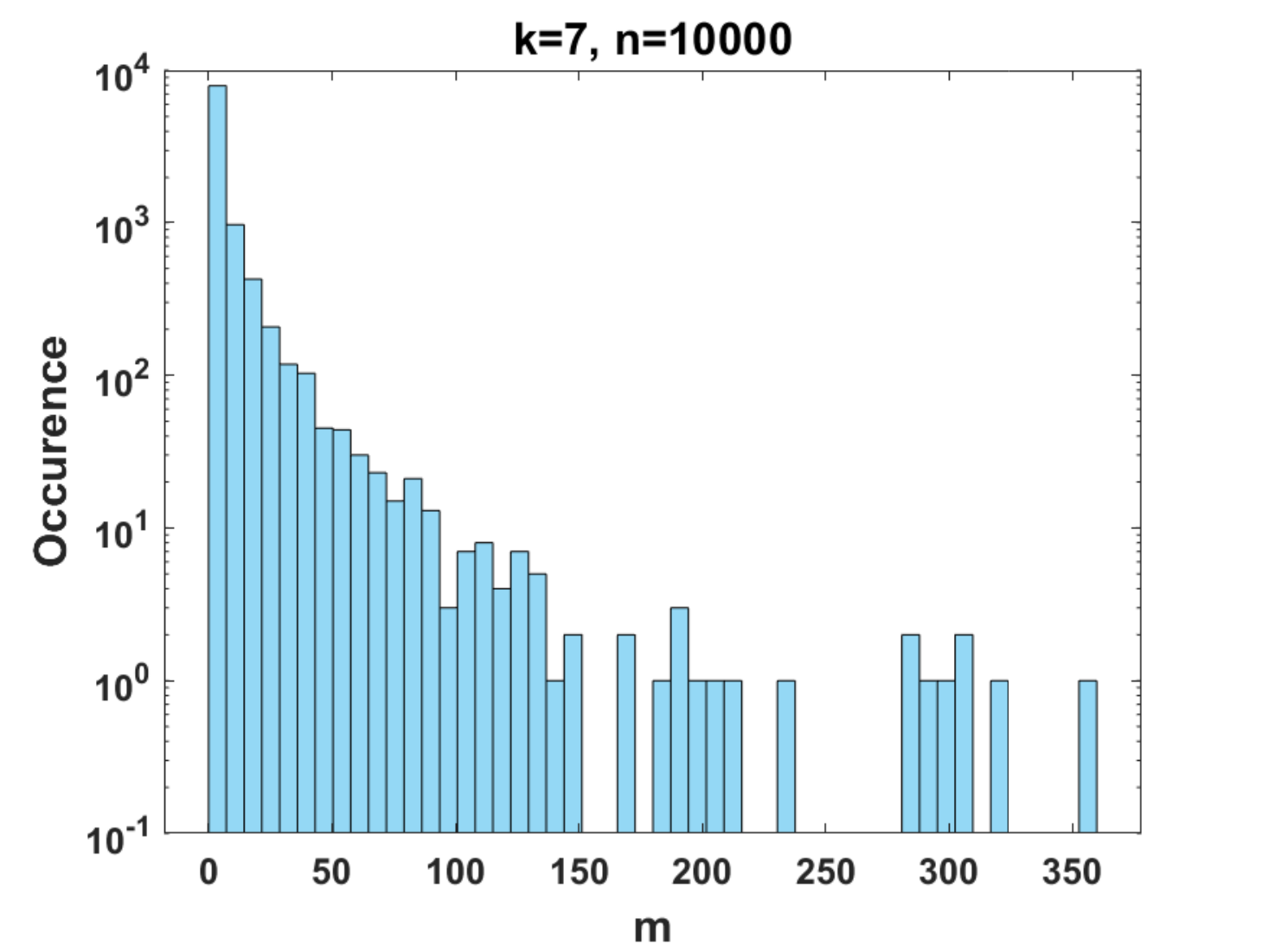}%
   \includegraphics[width=.32\linewidth]{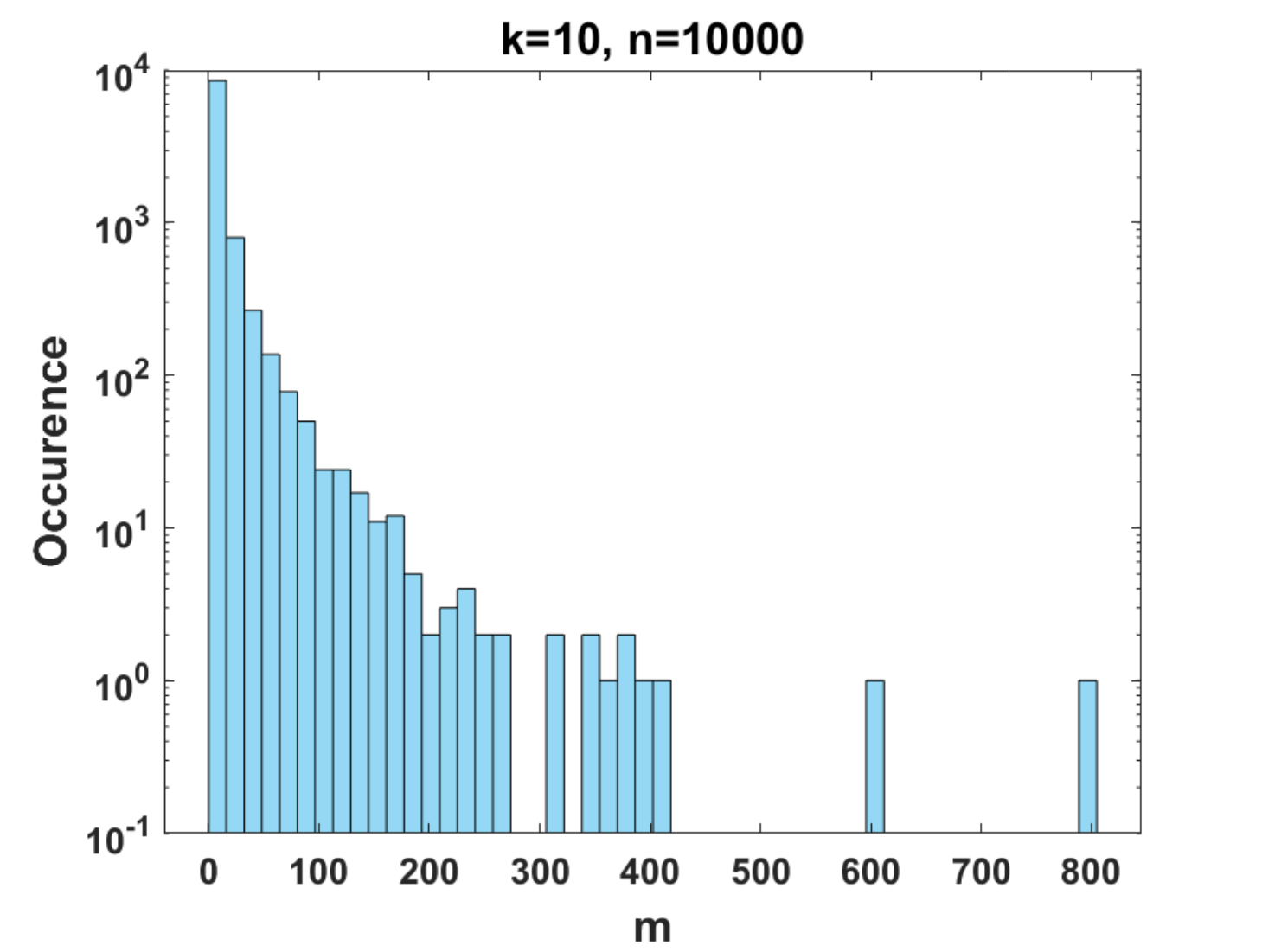}%
   \caption{BigGAN~\cite{brock2018large}, $Z$-space ($128$ dimensions)}
 \end{subfigure}
    \caption{
    Distributions of $m$-hub latents for state-of-the-art GANs, $k=5,7,10$ (the $k$-NN algorithm) and $n=10000$ (size of latent sample set $S$).
    All distributions are highly tailed to the right, which shows the existence of hubness phenomenon~\cite{radovanovic2010hubs} in GAN latent spaces. Similar phenomena hold for StyleGAN $Z$-space and ProGAN (Appendix~\ref{appendix:frequency_distribution_additional}). Note that $y$-axis is in log-scale.}
    \label{fig:exploration_hubness}
\end{figure*}

\section{Related Work}

\subsection{Generative Adversarial Network (GAN)} 
Since the seminal work of \citeauthor{goodfellow2014generative} (2014), Generative Adversarial Networks (GANs) have become a major type of deep generative models and have been extensively studied in recent years.
Existing works mostly focus on the choices of architectures~\cite{radford2015unsupervised,he2016deep,jiang2021transgan}, loss functions~\cite{arjovsky2017wasserstein,mao2017least}, regularization and normalization techniques~\cite{gulrajani2017improved,mescheder2018training,miyato2018spectral,qin2020does}, aiming to stabilize the training of GANs and improve the quality of synthesized images.
To date, the best-performing GANs include the ProGAN~\cite{karras2018progressive} and the StyleGAN series~\cite{karras2019style,karras2020analyzing,karras2021alias} developed by Nvidia for unconditional image synthesis, and the BigGAN~\cite{brock2018large} developed by DeepMind for conditional image synthesis.
Despite the success of these methods, they all have the long-standing problem that a large proportion of the images synthesized from randomly sampled latents are low-quality ones with artifacts, which hinders their applications in visual content generation.

\vspace{2mm}
\noindent \textbf{GAN Latent Sampling.} To sidestep the above-mentioned problem and obtain high-quality synthesized images, two workaround solutions were proposed: i) ``cherry-picking'' and ii) the truncation trick~\cite{marchesi2017megapixel,brock2018large,karras2019style}.
Between them, the first approach is a naive solution as one can always ``cherry-pick'' high-quality ones from a set of synthesized images in an {\it a posteriori} manner with visual inspection. Obviously, this method is inefficient as it requires intensive human labor and is not applicable for large-scale image synthesis tasks.
Unlike ``cherry-picking'', the truncation trick is an automatic method that can synthesize high-quality images by normalizing sampled latents to be close to their mean.
However, it is a purely {\it empirical} method with few insights.
In this paper, we propose a novel latent sampling method for GANs based on the observation of {\it hubness} phenomenon in their high dimensional latent spaces, which is efficient with solid theoretical insights.
We also show that the truncation trick is a naive approximation of our method due to the ``central clustering effect'' of {\it hub} latents. 

\subsection{Hubness Phenomenon}

{\it Hubness} is a well-known phenomenon that describes the impact of the notorious ``curse of dimensionality'' on nearest neighbors~\cite{radovanovic2010hubs}.
In a nutshell, the hubness phenomenon is proved to be an inherent property of data distributions in high-dimensional space~\cite{newman1983nearest,newman1985nearest,radovanovic2010hubs}, which uncovers an interesting but counter-intuitive fact: high dimensionality leads to the emergence of ``popular'' nearest neighbors ({\it a.k.a.} the {\it hub} points).
In other words, the hub points are those that are much more likely to be among the $k$-nearest neighbours of other points in a sample set. This fact poses challenges for algorithms that rely on nearest neighbor search.
Addressing such challenges, hubness-aware methods were proposed and applied in various areas, {\it e.g.} gene expression classification \cite{buza2016classification,buza2016semi}, time-series classification \cite{tomavsev2015hubness} and electroencephalograph classification \cite{buza2016ele}.
Meanwhile, hubness-aware $k$-nearest neighbor ($k$-NN) methods were also proposed, {\it e.g.} hubness-weighted $k$-NN \cite{radovanovic2010hubs}, hubness-fuzzy $k$-NN \cite{tomavsev2014hubness}, hubness-information $k$-NN \cite{tomasev2011nearest}, Naive Hubness-Bayesian $k$-NN \cite{tomavsev2011probabilistic}, and Augmented Naive Hubness-Bayesian $k$-NN \cite{tomavsev2013hub}.

In this paper, in contrast to previous methods that treat hubness as an undesirable phenomenon and aim to mitigate it, we show that the hubness phenomenon can be effectively used as priors for the sampling of high-quality GAN latents that produce high-quality synthesized images.

\begin{figure*}[htp]
\centering
    \begin{subfigure}[h]{0.32\textwidth}
      \centering
        \includegraphics[width=0.99\linewidth]{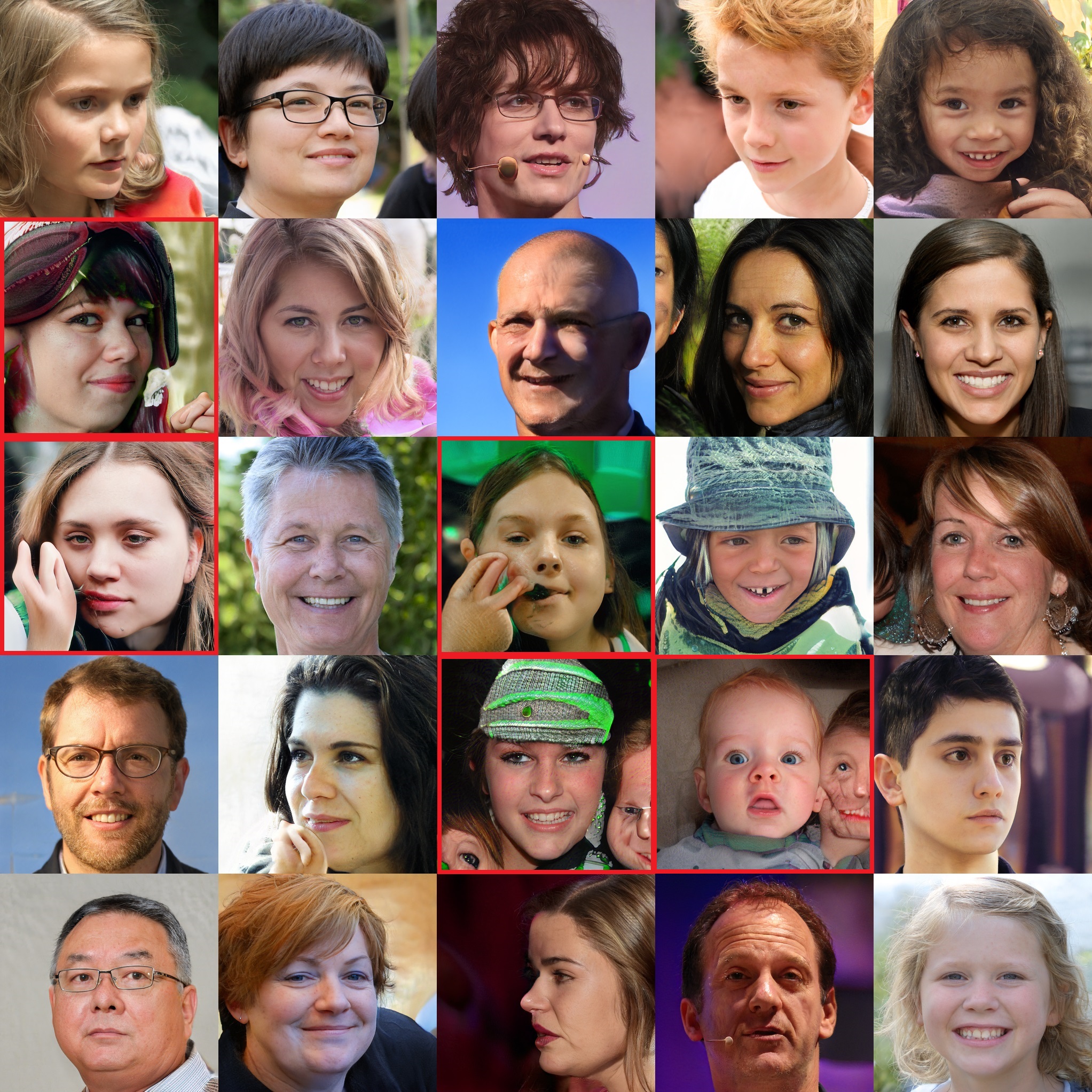}%
      \caption{Baseline (random latent sampling)}
    \end{subfigure}
    \begin{subfigure}[h]{0.32\textwidth}
        \label{fig:n10000_hubs}
        \centering
        \includegraphics[width=0.99\linewidth]{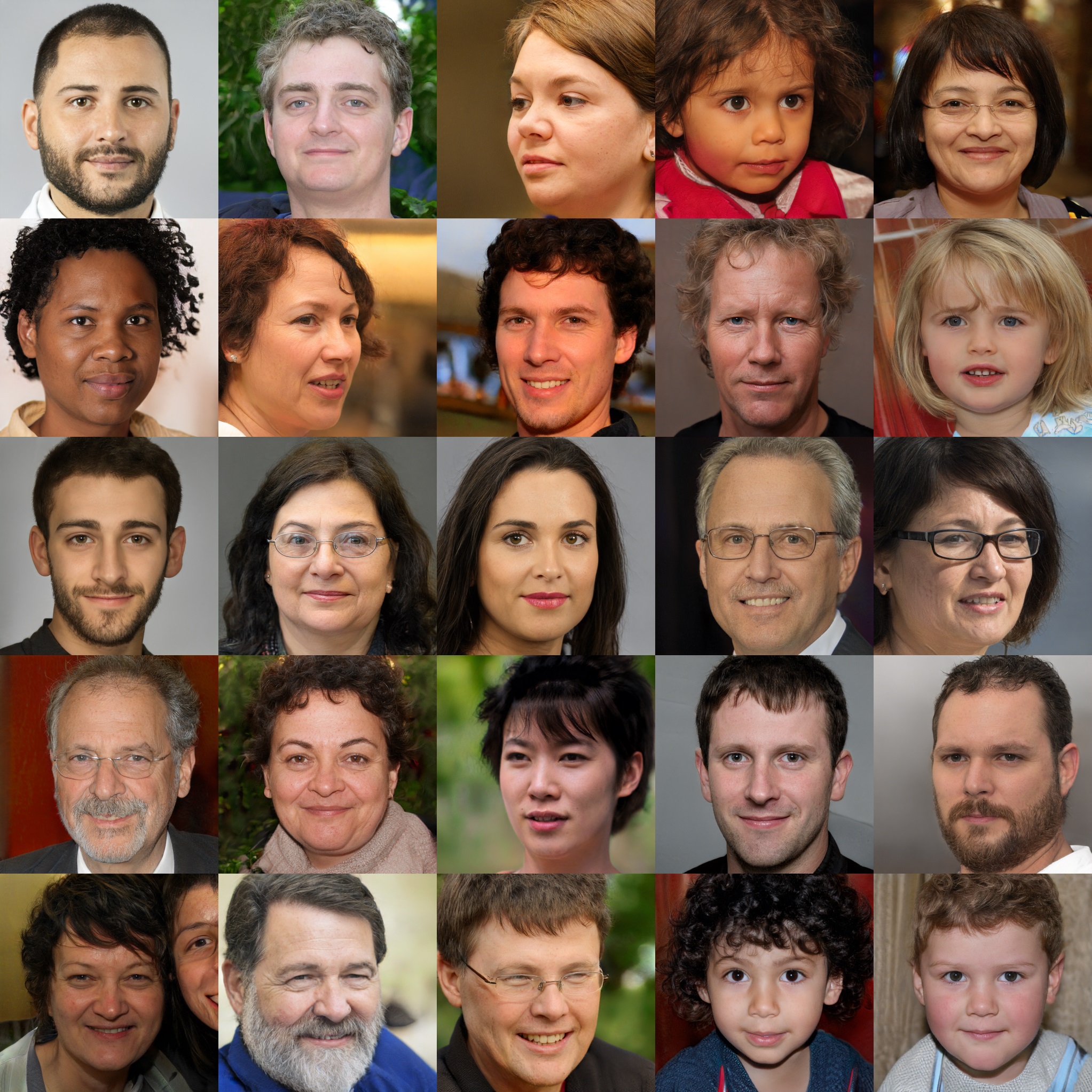}%
      \caption{Our method ($W$-space hubness priors)}
    \end{subfigure}
    \rulesep
    \begin{subfigure}[h]{0.32\textwidth}
       \includegraphics[width=.99\linewidth]{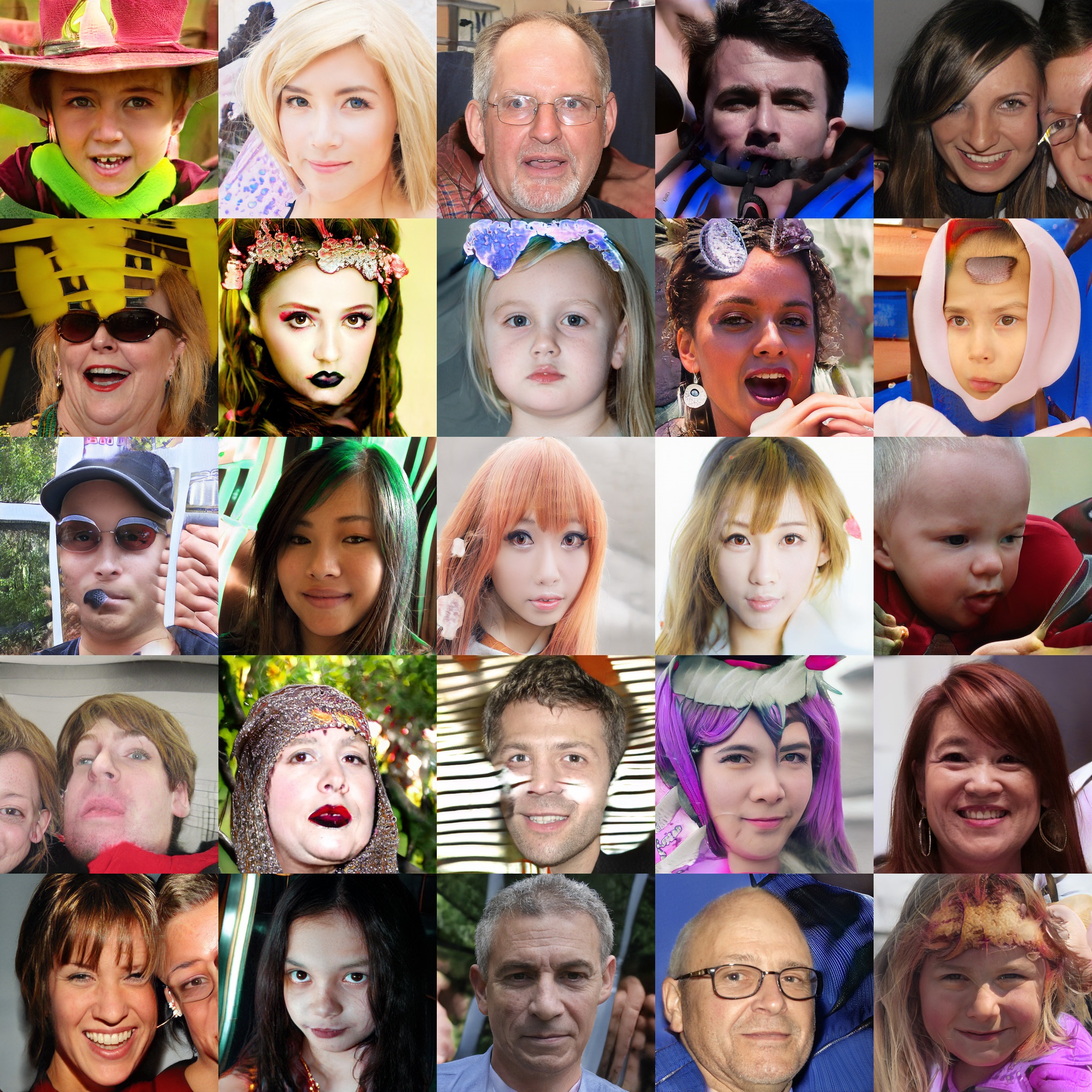}%
        \caption{Low-quality latent sampling (Alg.~\ref{alg:lq_sampling})}
    \end{subfigure}
    \caption{(a) and (b): Effectiveness of our method (hubness priors) against the baseline (random latent sampling). We use $n=10000$, $k=5$ and hub value threshold $t=50$ in our method. The StyleGAN2~\cite{karras2020analyzing} images generated using our method are almost always of high quality while those generated using the baseline contain both high-quality and low-quality (\textcolor{red}{red} boxes) results. (c): Low-quality StyleGAN2~\cite{karras2020analyzing} images generated using the {\bf reversed} version of our method, {\it i.e.} Algorithm~\ref{alg:lq_sampling} (Appendix~\ref{appendix:pseudocode_lq_sampling}), where $n=10000$, $k=5$ and hub value threshold $t_{lq}=1$. Almost all images are of low quality.}
    \vspace{-4mm}
    \label{fig:effectiveness}%
\end{figure*}

\section{Hubness Priors for GAN Latent Sampling}

In this section, we first explore the hubness of GAN latents (Section \ref{sec:exploration_hubness}) and then exploit the insights obtained as priors to develop a novel algorithm for the sampling of high-quality latents for GANs (Section \ref{sec:exploitation_hubness}).

\subsection{Exploring Hubness of GAN Latents}
\label{sec:exploration_hubness}

Inspired by previous studies on the hubness phenomenon of data distributions in high dimensional space \cite{radovanovic2010hubs}, let $Z \in \mathbb{R}^d$ be a $d$-dimensional GAN latent space, $S = \{z_1, z_2, ..., z_n\}$, $z_i \in Z$ be a set of latents sampled from a $d$-dimensional standard normal distribution $\mathcal{N}(0,I)$, $k$ be the parameter of the $k$-nearest neighbor algorithm, we define {\it $m$-hub latents} as:
\begin{definition}
Latent code $z_i$ ($1\leq i \leq n$) is an $m$-hub latent if $z_i$ is among the $k$-nearest neighbors of $m$ ($m < n$) sampled latents in $S$, where $m$ is the {\it hub value} of $z_i$.
\end{definition}
With the above definition, we explore the hubness of GAN latents by investigating the distributions of $m$-hub latents in the latent spaces of state-of-the-art GANs~\cite{brock2018large,karras2018progressive,karras2019style,karras2020analyzing,karras2021alias}. 
As Fig.~\ref{fig:exploration_hubness} shows, it can be observed that the distributions of $m$-hub latents are highly tailed to the right.
Thus, we argue that the GAN latents are not uniformly distributed and that a small portion of them are much more likely to be close to other latents in the latent space, {\it i.e.,} with large $m$.
Therefore, these latents tend to have larger sampling densities and are thus better trained than other latents during GAN training.
Based on the heuristics that well-trained latents are more likely to yield high-quality images, we conjecture that the hubness phenomenon can be used as priors to identify GAN latents that generate high-quality results:
\begin{conjecture}
\textbf{(Hubness Priors)}
The quality of GAN synthesized images and the hub values $m$ of their corresponding latents are positively correlated.
\label{conjecture:hubs_priors}
\end{conjecture}
Please see Section~\ref{sec:effHubsPriors} for an empirical justification of our conjecture.

\newlength{\textfloatsepsave} \setlength{\textfloatsepsave}{\textfloatsep}
\setlength{\textfloatsep}{10pt}
\begin{algorithm}[h]
\textbf{Input:} a set of GAN latents $S=\{z_1, z_2, ..., z_n\}$ randomly sampled from a standard normal distribution $\mathcal{N}(0,I)$, a hyper-parameter $k$, a threshold $t$\\
\textbf{Output: $S_{hq}$} 
\begin{algorithmic}
\caption{GAN Latent Sampling with Hubness Priors}
\label{alg:hq_sampling}

\STATE \# Step 1
\STATE $m_{1,2,...,n} \gets 0$

\FOR{$i \gets 1 $ to $n$}
    \STATE $\{\mathrm{idx}_1, \mathrm{idx}_2, ...\mathrm{idx}_k\} \gets$ $k$-NN($z_i$)
    \FOR{$j \gets 1 $ to $k$}
        \STATE $m_{\mathrm{idx}_j} \gets m_{\mathrm{idx}_j} + 1$
    \ENDFOR
\ENDFOR
\STATE \# Step 2
\STATE $S_{hq} \gets \emptyset$
\FOR{$i \gets 1 $ to $n$}
    \IF{$m_i \geq t$}
        \STATE  $S_{hq} \gets S_{hq} \cup z_{i}$
    \ENDIF
\ENDFOR
\end{algorithmic}
\end{algorithm}

\vspace{2mm}
\noindent \textbf{Remark on Random Latent Sampling}
Previously, it was widely believed that GAN latents are {\it unbiased} as they are sampled from a simple but well-behaved noise distribution, {\it i.e.,} the standard normal distribution  $\mathcal{N}(0,I)$, which is isotropic and has most of its density on a hypersphere surface in the high-dimensional GAN latent space \cite{menon2020pulse}.
This implies that all sampled latents should be approximately uniformly distributed and of similar norms\footnote{In latest implementations~\cite{karras2019style,karras2020analyzing,karras2021alias}, the latents are explicitly normalized to be of the same norm.}, thereby contributing to the sampling in a similar manner.
While in this paper, we counter this popular belief by showing that GAN latents are actually {\it biased} from the observation of hubness phenomenon in GAN latent spaces.
Among all latents, the {\it hub} ones tend to have higher sampling densities and are thus better trained by GANs, thereby generating higher quality images.

\setlength{\textfloatsep}{\textfloatsepsave}
\begin{figure}[t]
\centering
    \centering
    \subfloat[\centering ProGAN-HQ ]{\includegraphics[width=.42\linewidth]{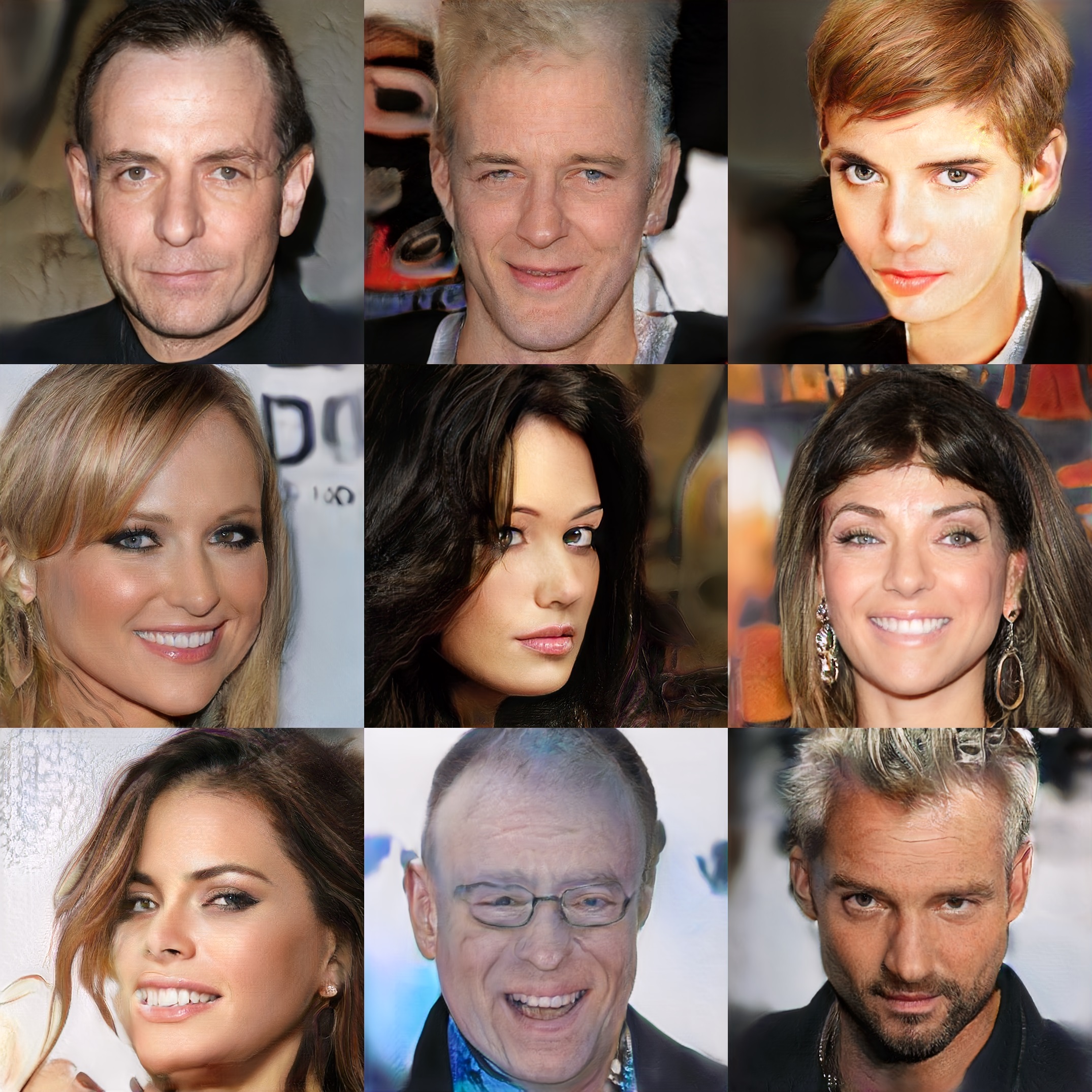}}%
    \qquad
    \subfloat[\centering ProGAN-LQ ]{\includegraphics[width=.42\linewidth]{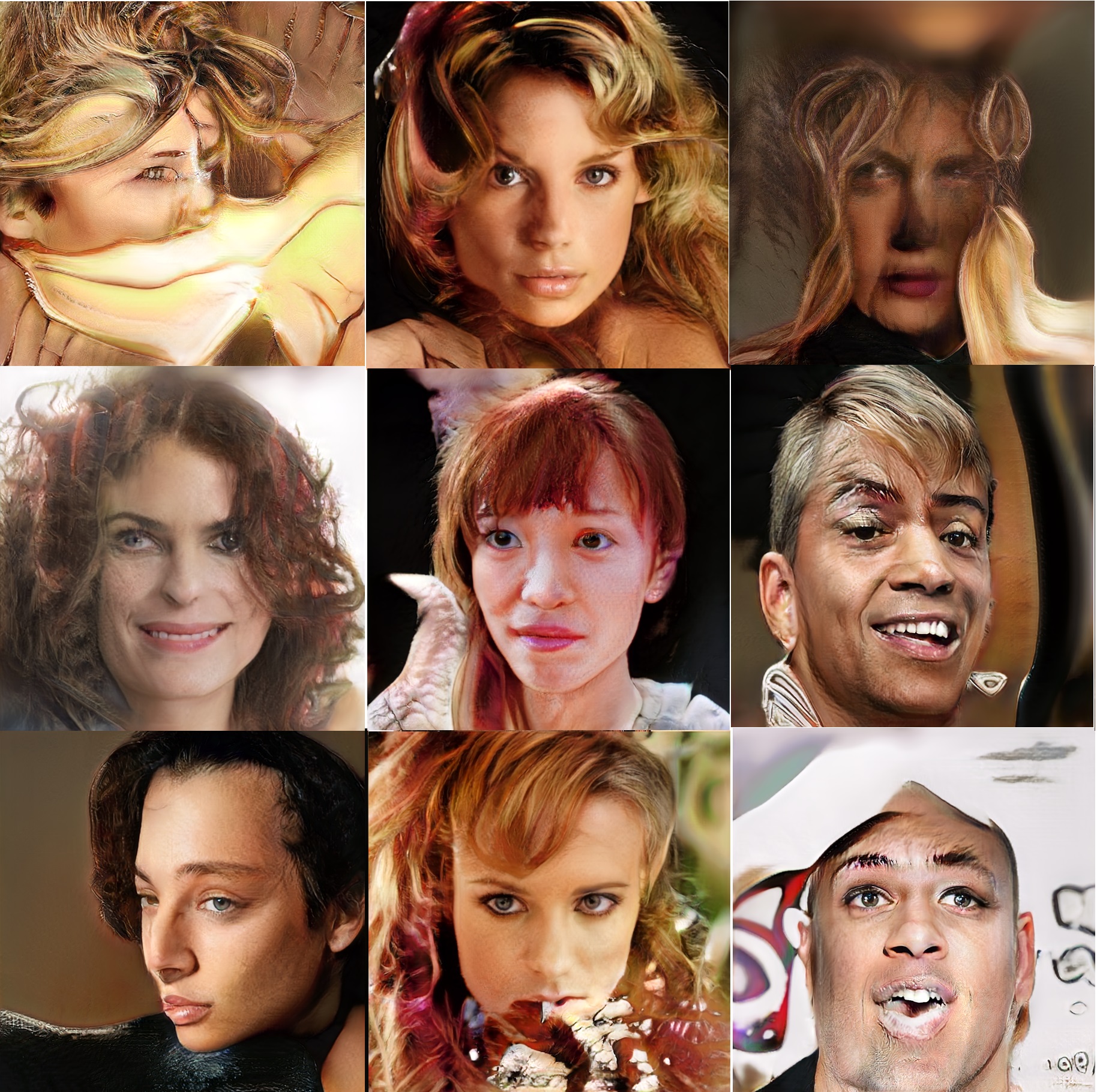}}
    \qquad
    \subfloat[\centering BigGAN-HQ ]{\includegraphics[width=.42\linewidth]{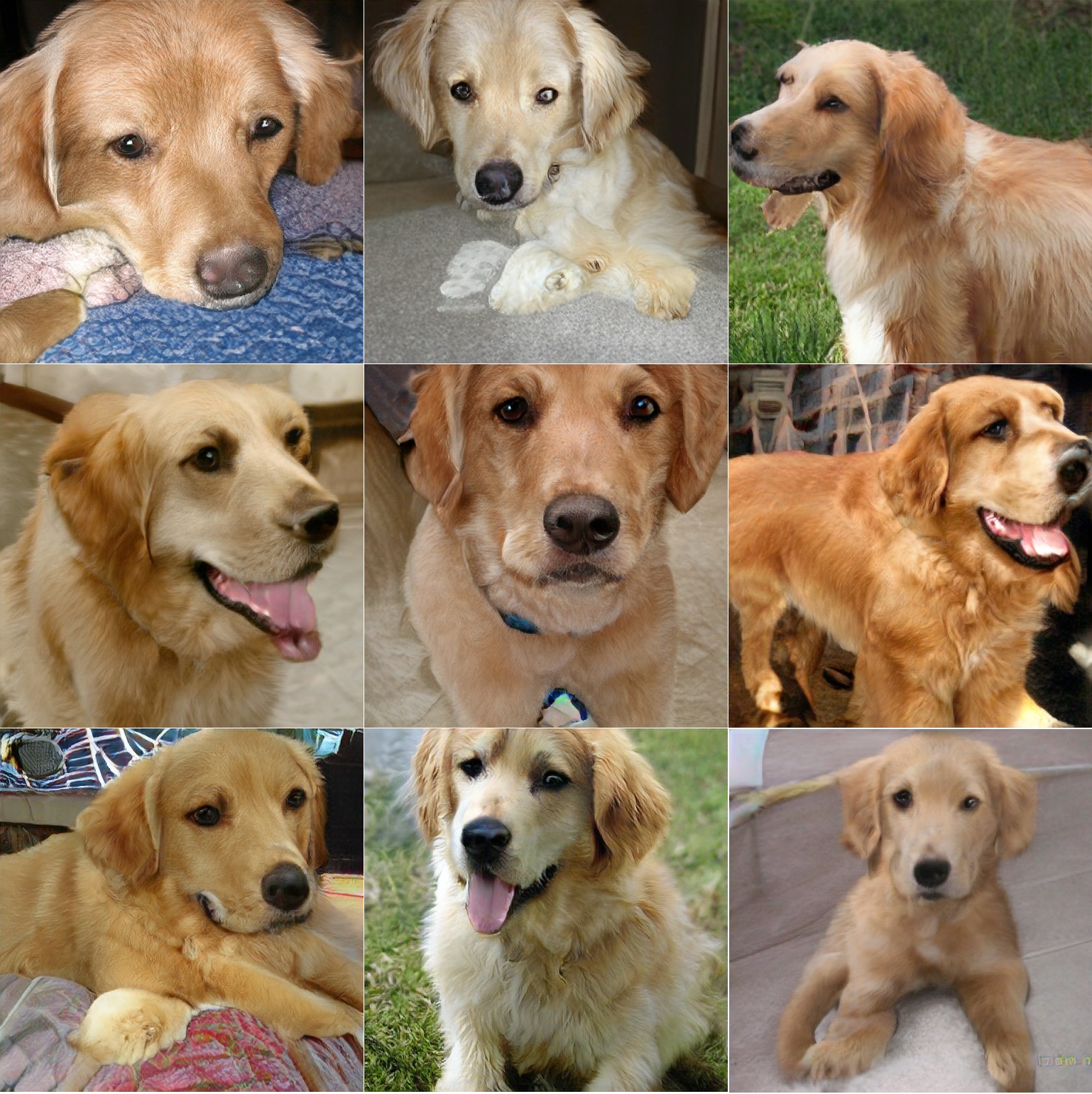}}%
    \qquad
    \subfloat[\centering BigGAN-LQ]{\includegraphics[width=.42\linewidth]{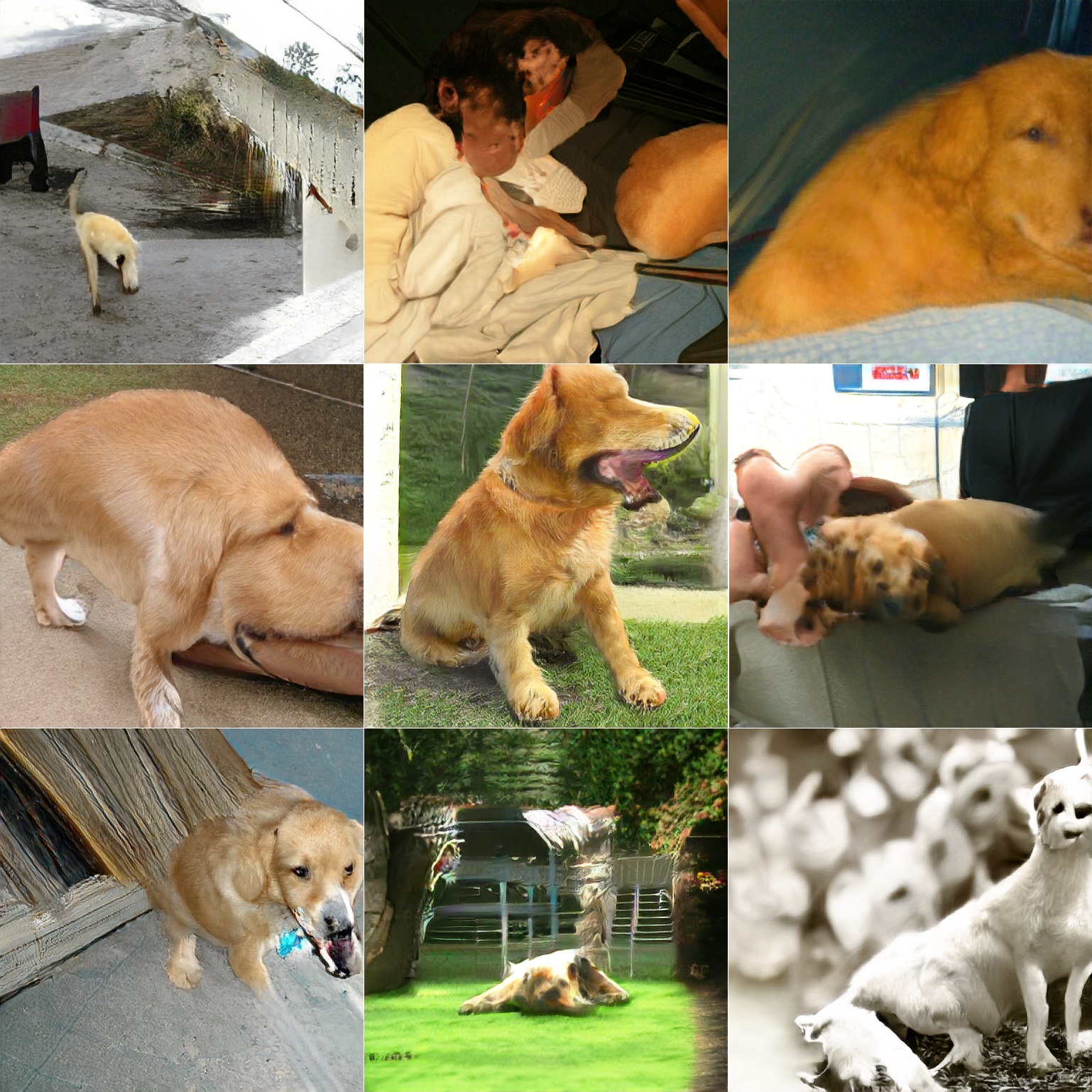}}%
    \qquad
    \subfloat[\centering StyleGAN3-HQ ]{\includegraphics[width=.42\linewidth]{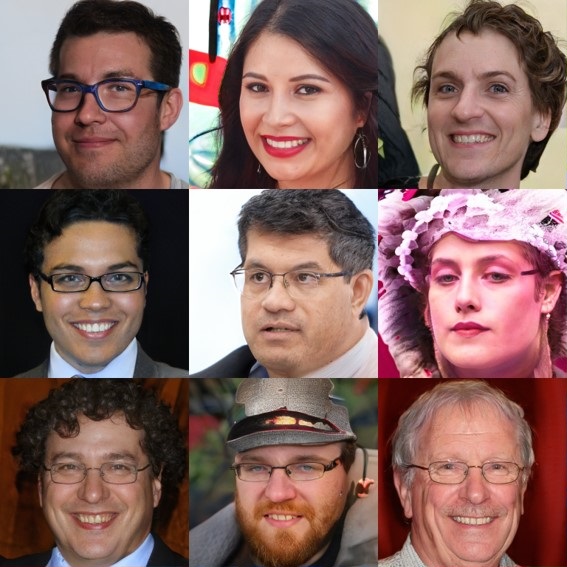}}%
    \qquad
    \subfloat[\centering StyleGAN3-LQ]{\includegraphics[width=.42\linewidth]{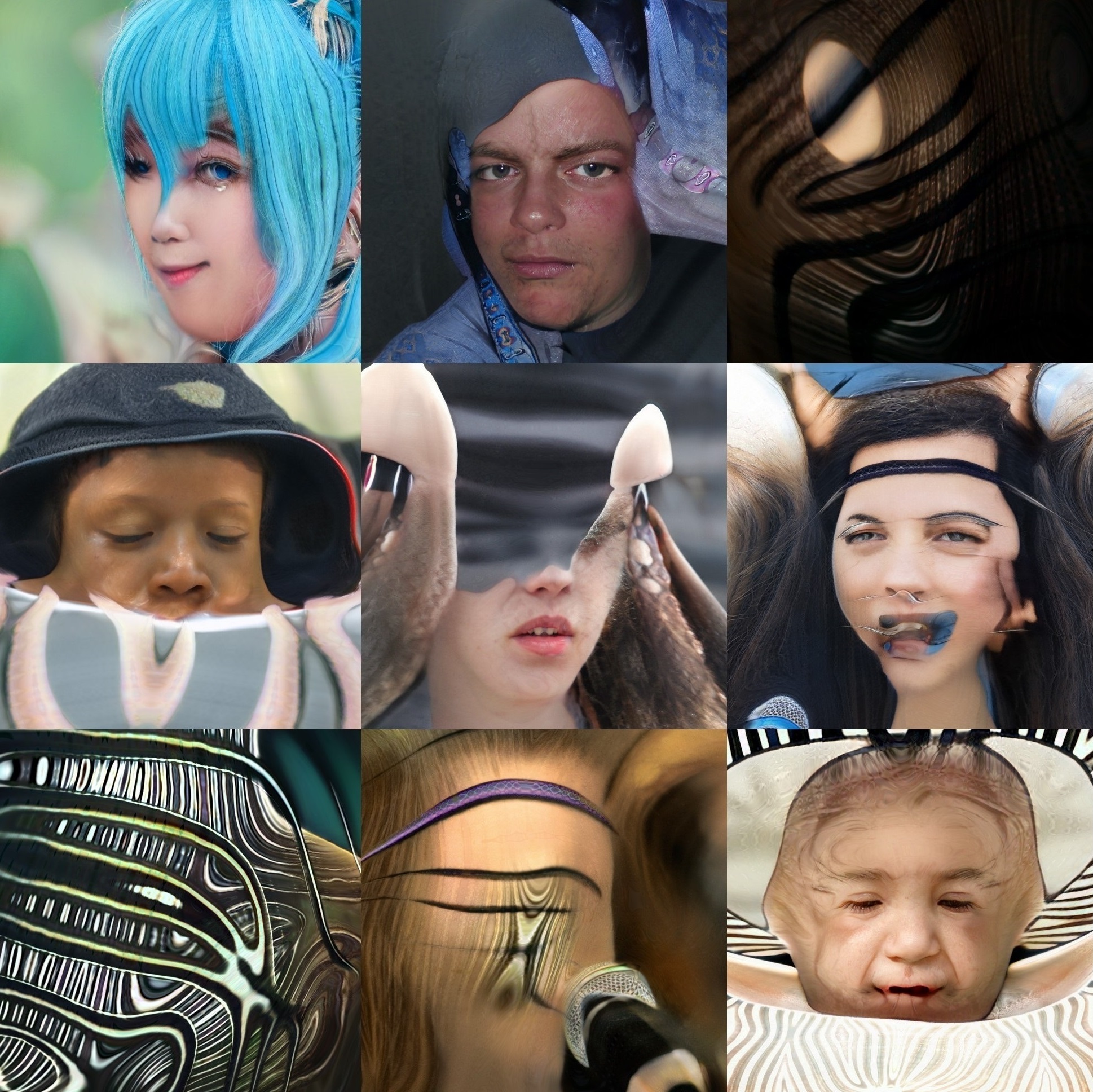}}%
    \caption{Performance of our method on ProGAN~\cite{karras2018progressive}, BigGAN~\cite{brock2018large} and StyleGAN3~\cite{karras2021alias}. It can be observed that our method works well on all GAN architectures.
    (a) and (b), (c) and (d), (e) and (f) are images synthesized using high-quality (HQ) and low-quality (LQ) latents obtained by our method with ProGAN, BigGAN and StyleGAN3 respectively. We use Algorithm~\ref{alg:hq_sampling} to obtain HQ latents and Algorithm~\ref{alg:lq_sampling} (Appendix~\ref{appendix:pseudocode_lq_sampling}) to obtain LQ latents respectively. We use $n=10000$, $k=5$, $t=50$ and $t_{lq}=1$.}%
    \vspace{-6mm}
    \label{fig:otherGAN}%
\end{figure}
\begin{figure}[t]
\centering
    \centering
    \subfloat[\centering StyleGAN-Car-HQ ]{\includegraphics[width=.45\linewidth]{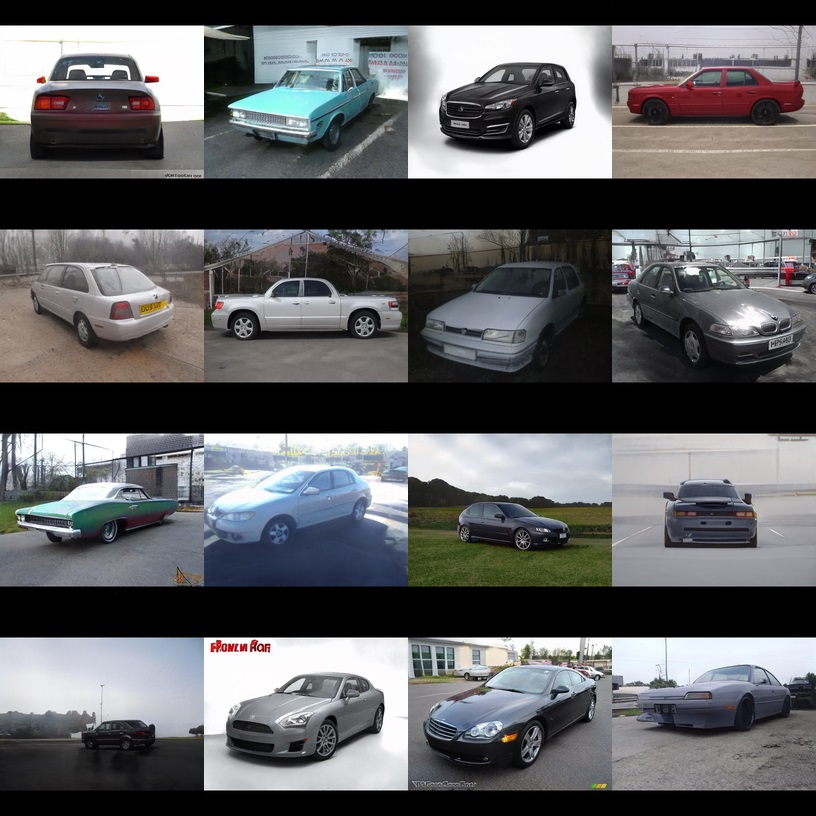}}%
    \qquad
    \subfloat[\centering StyleGAN-Car-LQ ]{\includegraphics[width=.45\linewidth]{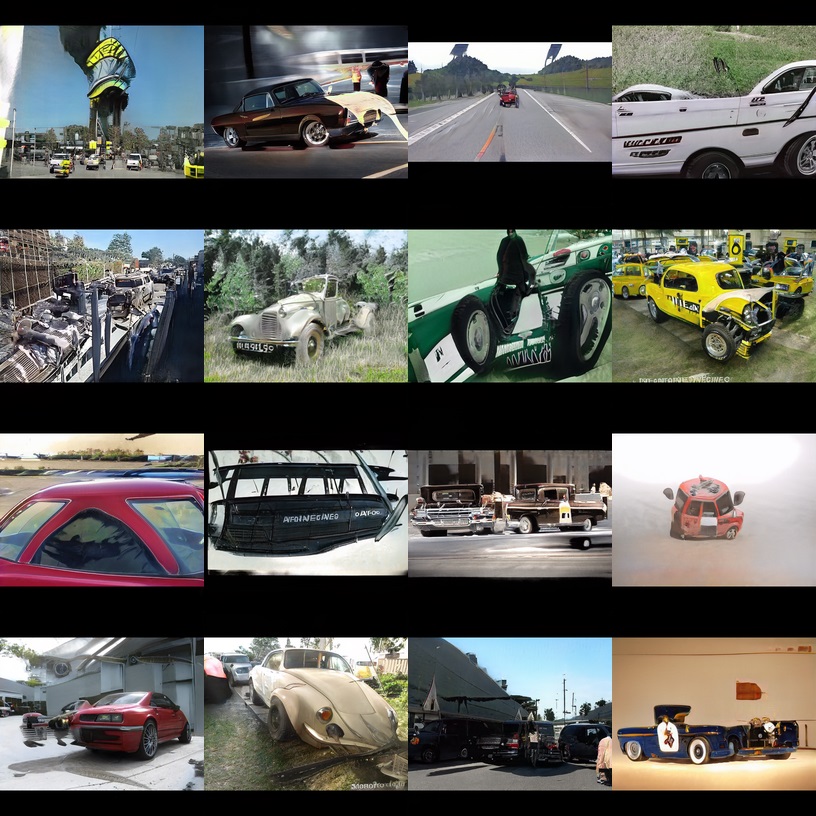}}
    \qquad
    \subfloat[\centering StyleGAN-Cat-HQ ]{\includegraphics[width=.45\linewidth]{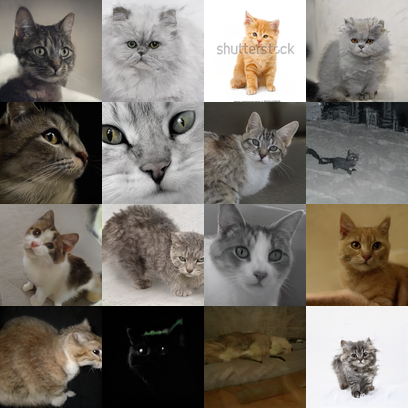}}%
    \qquad
    \subfloat[\centering StyleGAN-Cat-LQ
    ]{\includegraphics[width=.45\linewidth]{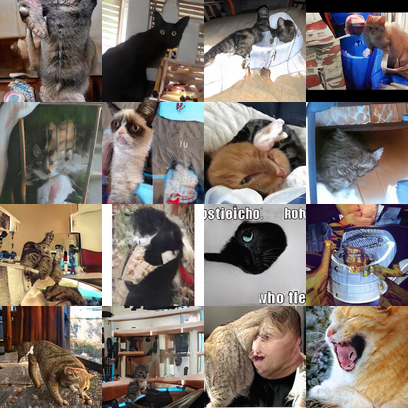}}%
    
    \caption{Performance of our method on StyleGAN2 pretrained on different image domains. It can be observed that our method works well on all domains.
    (a) and (b), (c) and (d) are images synthesized from high-quality (HQ) and low-quality (LQ) latents obtained by our method using a StyleGAN2 pretrained on the cars domain and the cats domain, respectively. We use Algorithm~\ref{alg:hq_sampling} to obtain HQ latents and Algorithm~\ref{alg:lq_sampling} (Appendix~\ref{appendix:pseudocode_lq_sampling}) to obtain LQ latents respectively. We use $n=10000$, $k=5$, $t=50$ and $t_{lq}=1$. Please see Appendix~\ref{appendix:frequency_distribution_additional} for results on the horse domain.}%
    \vspace{-6mm}
    \label{fig:otherdataset}%
\end{figure}

\subsection{Exploiting Hubness Priors for High-quality GAN Latent Sampling}
\label{sec:exploitation_hubness}

As Conjecture~\ref{conjecture:hubs_priors} states, the identification of high-quality GAN latents relies on their hub values $m$.
Thus, given a set of GAN latents $S=\{z_1, z_2, ..., z_n\}$ randomly sampled from a standard normal distribution $\mathcal{N}(0,I)$, a hyper-parameter $k$, and a threshold $t$, we utilize the proposed hubness priors and design a simple two-step GAN latent sampling algorithm:
First, we compute the hub value $m_i$ for each latent $z_i \in S$ using a standard $k$-NN ($k$-nearest neighbor) algorithm; 
Second, we identify $z_i$ as a high-quality latent if $m_i$ is larger than a user-defined threshold $t$, and add $z_i$ into a set $S_{hq}$.
The set $S_{hq}$ is the output of our algorithm, which contains all the high-quality latents identified.
Algorithm~\ref{alg:hq_sampling} shows the pseudocode of our algorithm.
Note that our algorithm is fundamental and widely applicable to different types of GANs as long as they sample latents from a standard normal distribution, {\it e.g.} conditional GANs~\cite{brock2018large}.

\noindent \textbf{Relationship to Truncation Trick.}
To our knowledge, the truncation trick ~\cite{marchesi2017megapixel,brock2018large,karras2019style} is the only a priori method to sample high-quality GAN latents before our work, which is based on a heuristic that high-quality latents are those close to their mean. However, such a heuristic is purely empirical with few insights. 
Surprisingly, the proposed {\it hubness priors} have revealed the rationale of the truncation trick: the {\it hub} latents obtained by our method tend to cluster towards their mean~\cite{radovanovic2010hubs}.
Thus, we argue that the well-known truncation trick is a naive approximation of our method as it only captures near-mean {\it hub} latents but overlooks those that are relatively far from the mean.
Please see Section~\ref{sec:comparison_truncation_trick} for an empirical justification of our claims.

\begin{figure}[t]
\centering
    \centering
    \subfloat[\centering StyleGAN-Z-HQ ]{\includegraphics[width=.45\linewidth]{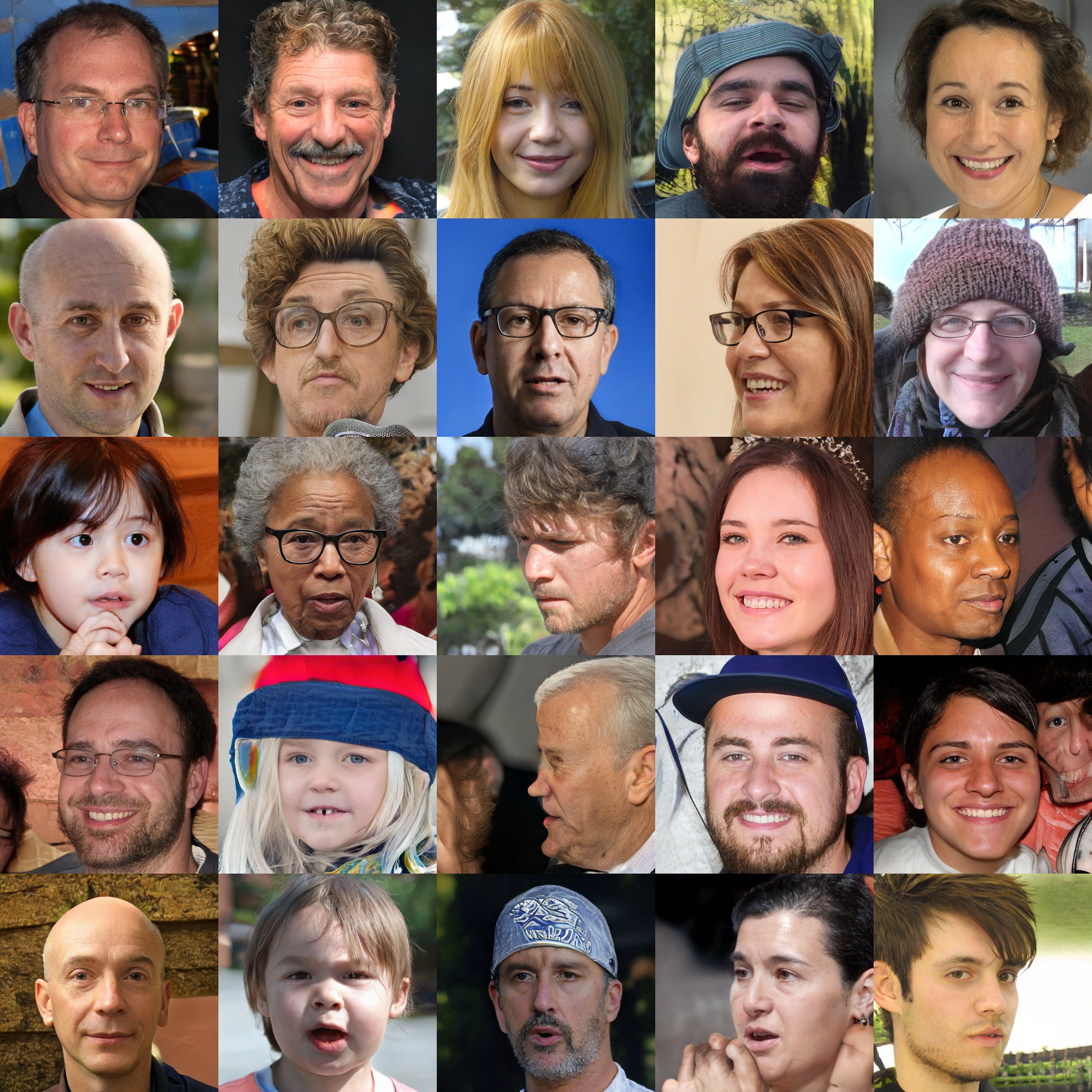}}%
    \qquad
    \subfloat[\centering StyleGAN-Z-LQ ]{\includegraphics[width=.45\linewidth]{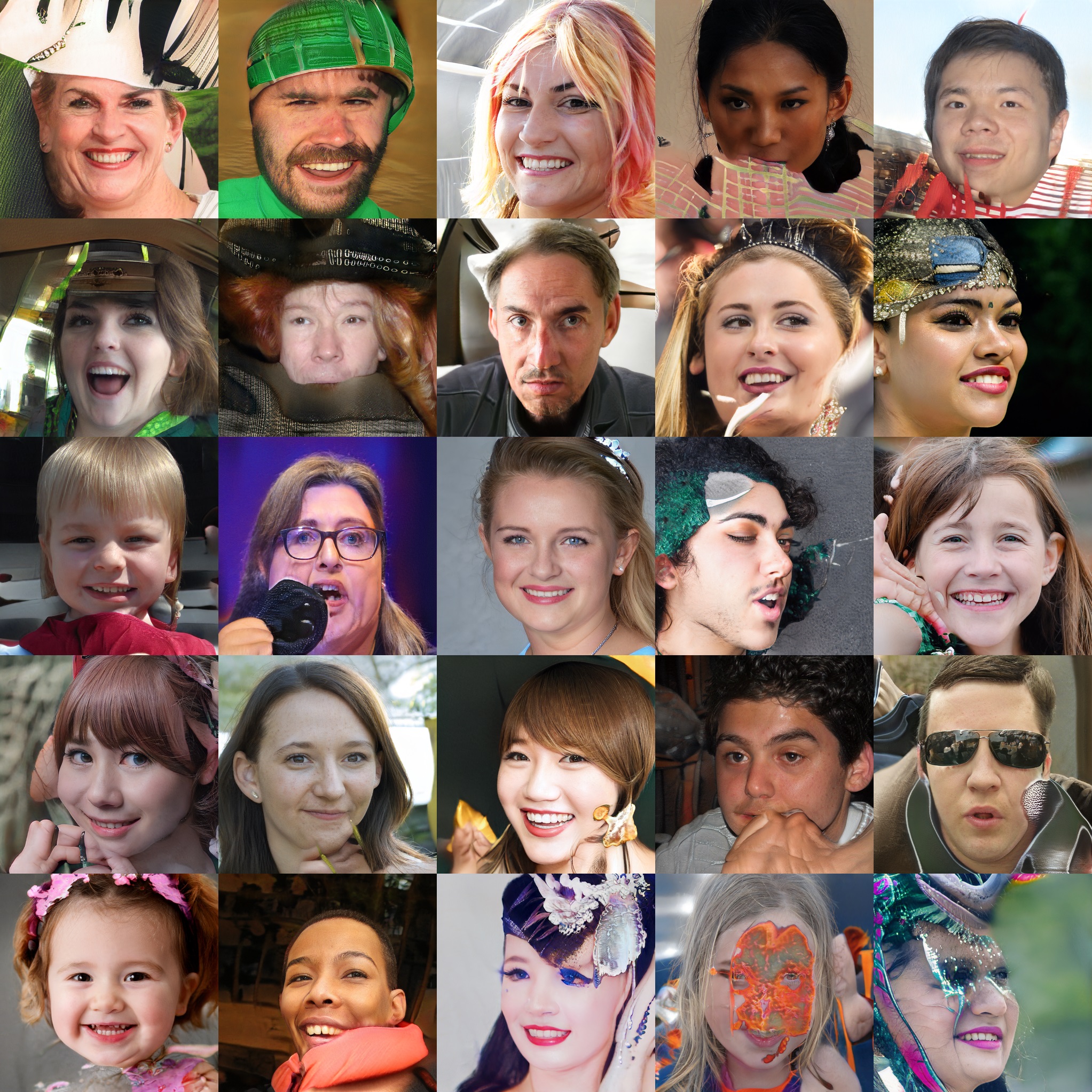}}
    \caption{Performance of our method on StyleGAN2's $Z$-space. We use $n=10000$, $k=5$, $t=50$ and $t_{lq}=1$.}%
    \vspace{-6mm}
    \label{fig:StyleGAN_z_space}%
\end{figure}

\section{Experimental Results}

\subsection{Experimental Setup}

Due to its a priori nature, our method allows for the sampling of high-quality GAN latents before the synthesis of images.
Thus, for the sampling of StyleGAN's $Z$-space and other GANs' latents, we use an Intel(R) Core(TM) i7-10875H CPU; for the sampling of StyleGAN's $W$-space latents, we use a GeForce RTX 2080 Ti GPU as the computation involves passing $Z$-space latents through a fully-connected mapping network~\cite{karras2019style,karras2020analyzing,karras2021alias}.
For the synthesis of high-quality images, we use publicly-released Github codes of StyleGANs\footnote{StyleGAN2,3: \url{https://github.com/NVlabs/stylegan2}, \url{https://github.com/NVlabs/stylegan3}.}~\cite{karras2019style,karras2020analyzing,karras2021alias}, BigGAN\footnote{ \url{https://github.com/ajbrock/BigGAN-PyTorch}}~\cite{brock2018large}, ProGAN\footnote{ \url{https://github.com/tkarras/progressive_growing_of_gans}}~\cite{karras2018progressive} with a GeForce RTX 2080 Ti GPU.
Unless specified, all results are generated with the $W$-space of StyleGAN2~\cite{karras2020analyzing}.
All quantitative results are averaged over three runs.
\textbf{Note that JPEG is applied to compress the synthesized images to meet the size limit. Please see the supplementary materials for uncompressed images.}

\subsection{Effectiveness of Hubs Priors}
\label{sec:effHubsPriors}

As Figs.~\ref{fig:effectiveness} (a) and (b) show, we compare the images generated by StyleGAN2~\cite{karras2020analyzing} using our method with those generated using the baseline, {\it i.e.} random latent sampling.
It can be observed that our method consistently yields high-quality images while the baseline generates both high-quality and low-quality images, which demonstrates the effectiveness of the proposed {\it hubness priors}.
Quantitatively, we observed better FID scores of images generated using our method than those by the baseline (Table~\ref{tab:FIDandDist}).

\vspace*{2mm}
\noindent \textbf{Low-quality Latents.}
As Conjecture~\ref{conjecture:hubs_priors} implies, the proposed {\it hubness priors} can also be used to identify low-quality latents that yield unrealistic synthesized images. Thus, as a complement to high-quality latent sampling, we implement low-quality GAN latent sampling by reversing the thresholding scheme in Algorithm~\ref{alg:hq_sampling} to $m_i \leq t_{lq}$ and have Algorithm~\ref{alg:lq_sampling} (Appendix~\ref{appendix:pseudocode_lq_sampling}).
As Fig.~\ref{fig:effectiveness} (c) shows, almost all synthesized images are of low quality, which justifies the effectiveness of the proposed {\it hubness priors}.

In fact, our {\it hubness priors} can be used to sort all sampled latents into a \textbf{\textit{hubness spectrum}} according to their hub values $m$ (Appendix~\ref{appendix:hubs_spectrum}), where the quality of images changes from high to low from left to right with decreasing $m$.

\subsection{Versatility}
\label{sec:versatility}

To demonstrate the versatility of our method, we show that it generalizes across different GAN architectures, different image domains and different latent spaces of the StyleGAN series~\cite{karras2019style,karras2020analyzing,karras2021alias}.

\vspace{2mm}
\noindent \textbf{Different GAN Architectures.} As Fig.~\ref{fig:otherGAN} shows, to justify that our method works across different GAN architectures, we show that our method also works on three other state-of-the-art GAN architectures, {\it i.e.} ProGAN~\cite{karras2018progressive}, BigGAN~\cite{brock2018large}, and the recent StyleGAN3~\cite{karras2021alias}.

\vspace{2mm}
\noindent \textbf{Different Image Domains.} 
As Fig.~\ref{fig:otherdataset} shows, to justify that our method works across different image domains, we show that our method also works on StyleGAN2 models pretrained on other images domains\footnote{All pre-trained networks are available at: \url{https://github.com/NVlabs/stylegan2}.}: cars, cats and horses.

\vspace{2mm}
\noindent \textbf{StyleGAN's $Z$-space.}
As Fig.~\ref{fig:StyleGAN_z_space} shows, our method also works for the $Z$-space of StyleGAN2~\cite{karras2020analyzing}. However, we observed that the quality variance of synthesized images is slightly lower when using the $W$-space. Thus, we propose to use the $W$-space for StyleGAN2.

\begin{figure*}[t]
\centering
    \centering
    \subfloat[\centering $t = 60$]{\includegraphics[width=.24\linewidth,valign=t]{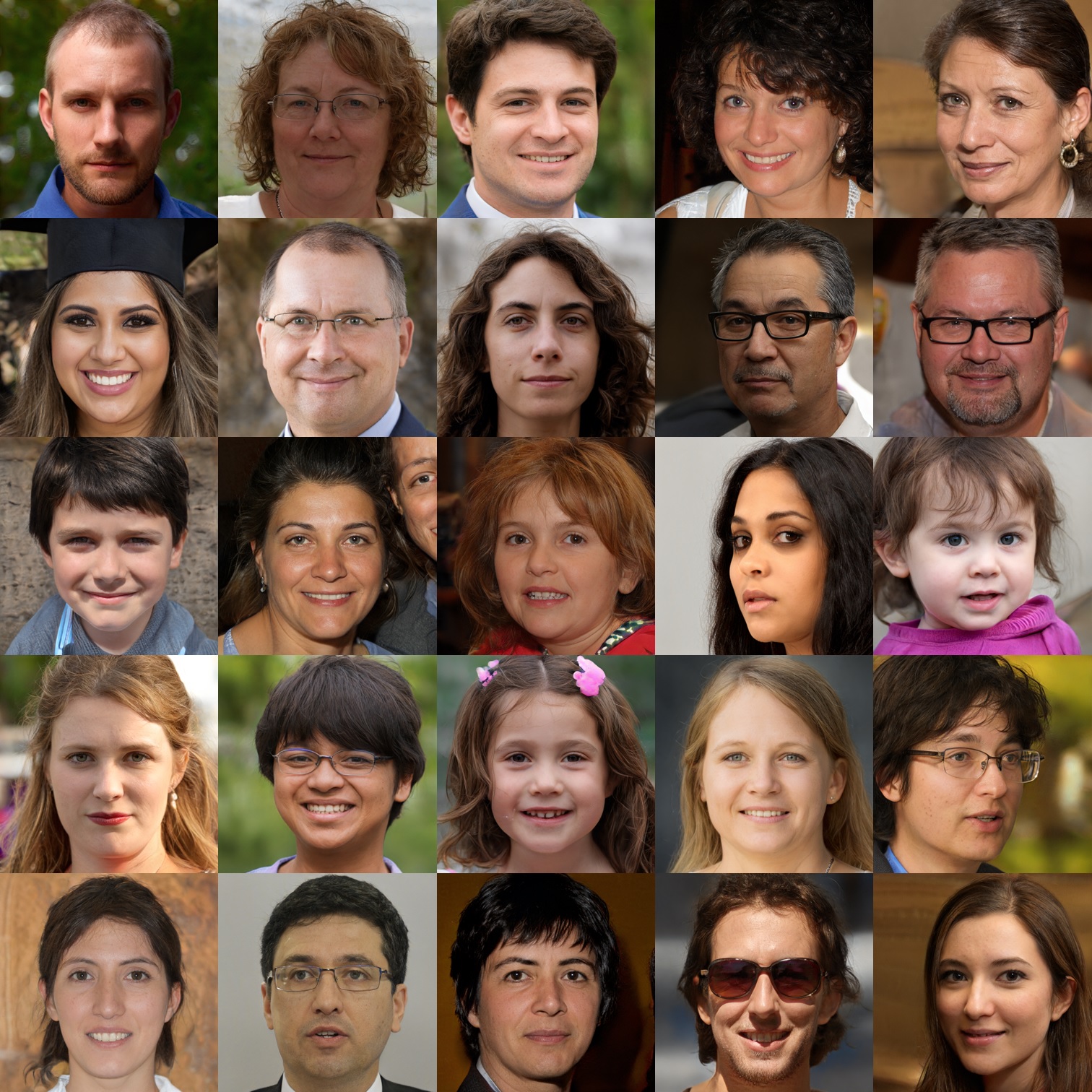}}%
    \,
    \subfloat[\centering $t = 50$]{\includegraphics[width=.24\linewidth,valign=t]{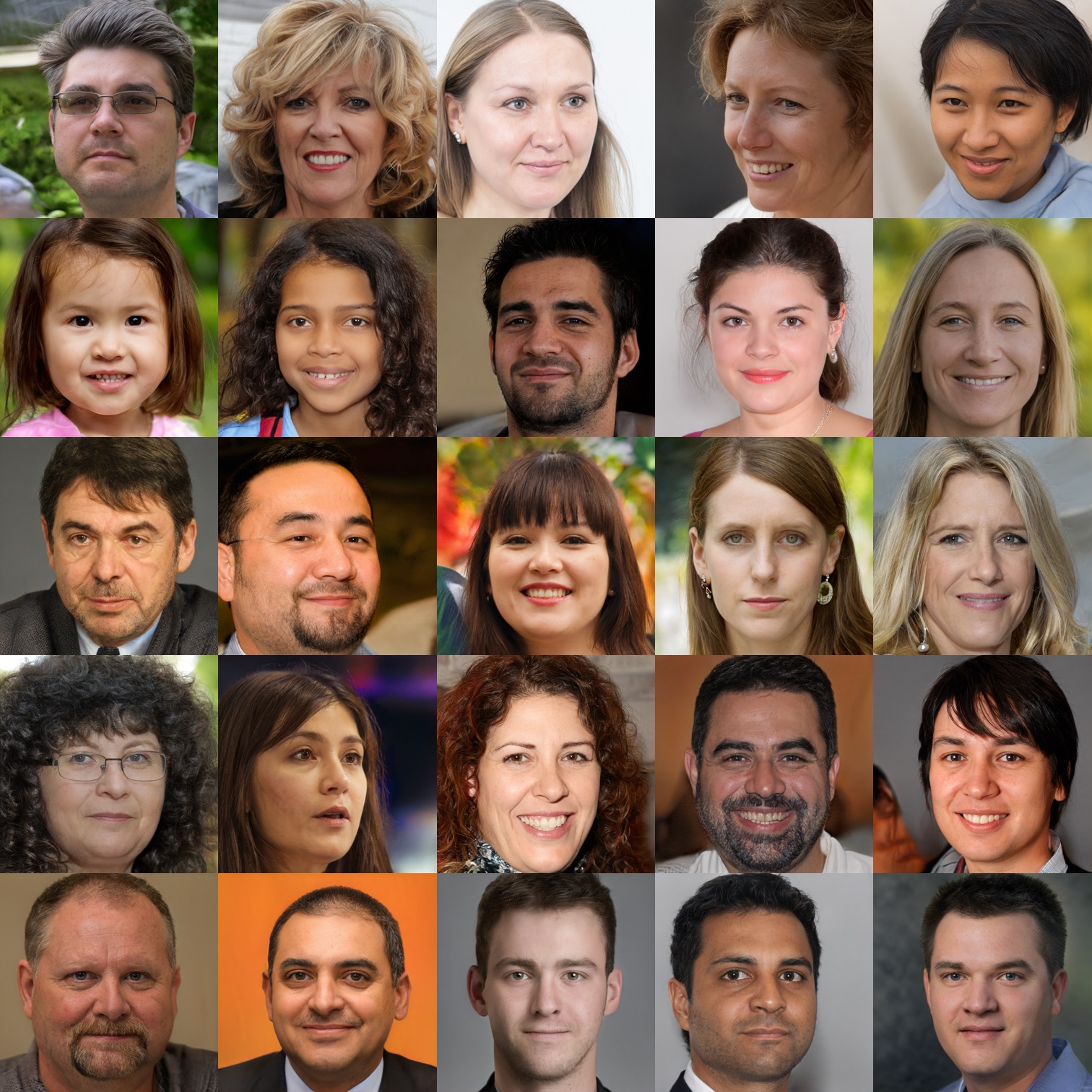}}%
    \,
    \subfloat[\centering $t = 40$]{\includegraphics[width=.24\linewidth,valign=t]{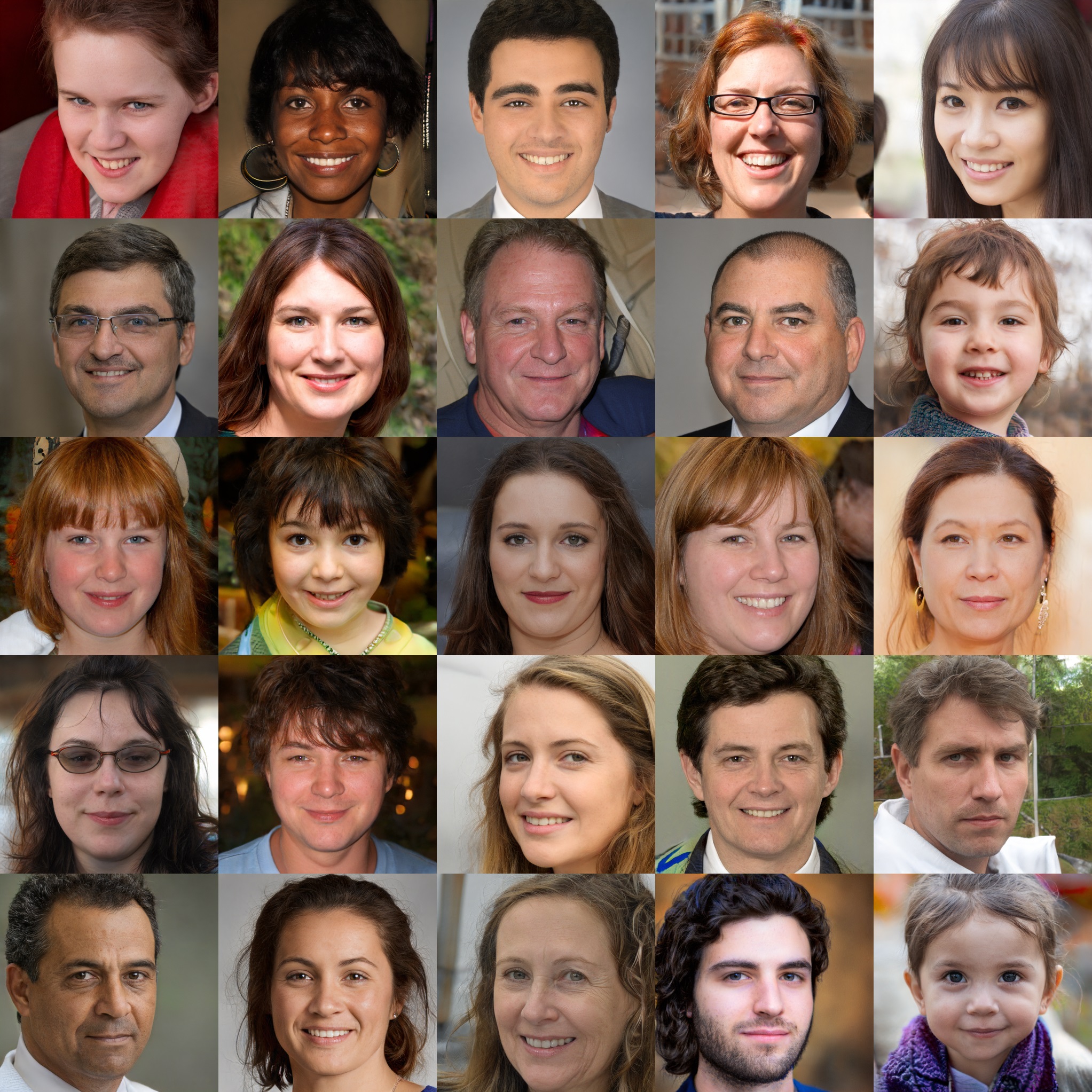}}
    \,
    \subfloat[\centering $t = 35$ ]{\includegraphics[width=.24\linewidth,valign=t]{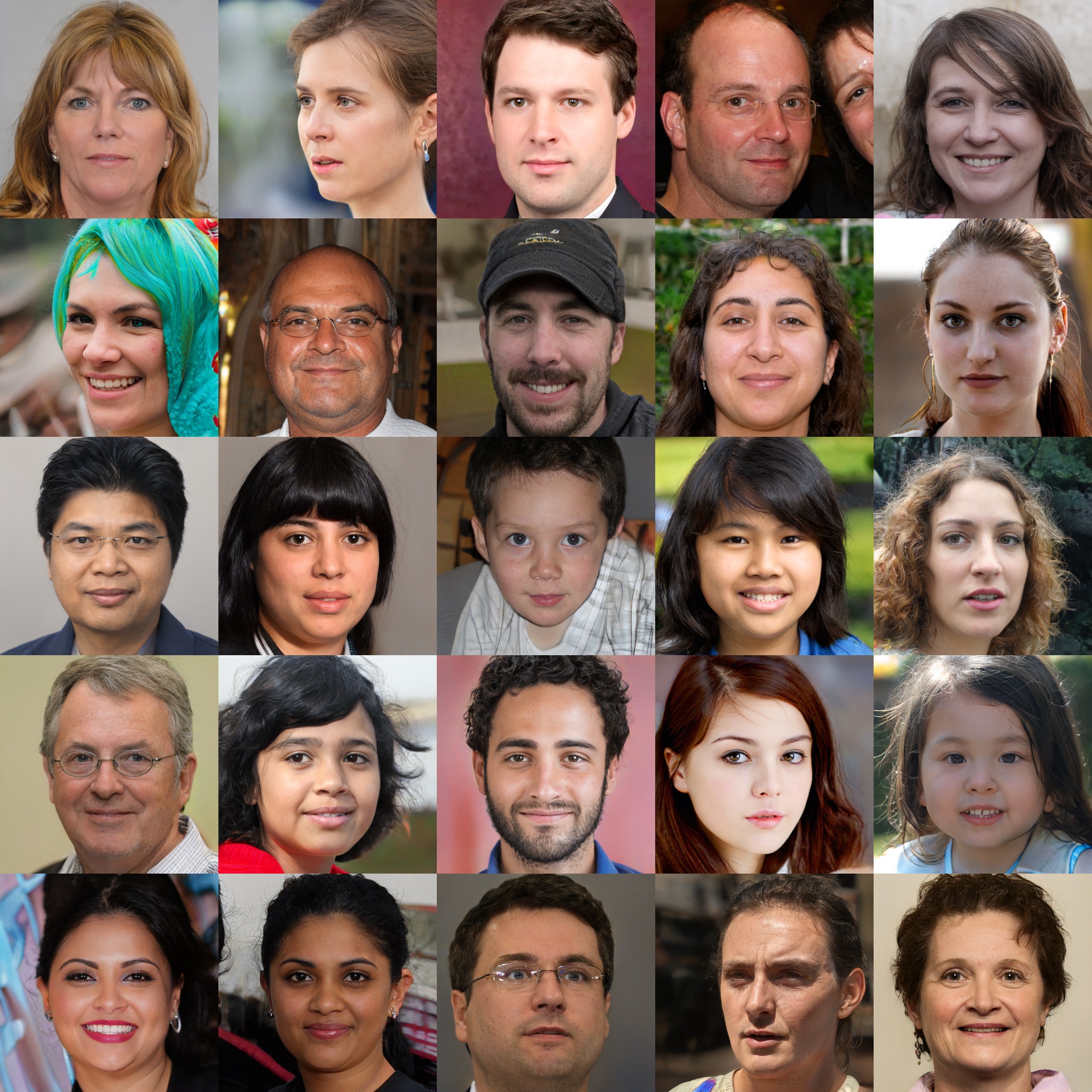}}%
    \caption{Performance of our method with different choices of threshold $t=60,50,40,35$. We use $n=10000$, $k=5$.}%
    \label{fig:diffTreshold}%
\end{figure*}
\begin{figure*}[t]
\centering
    \centering
    \subfloat[\centering $k = 3$ ]{\includegraphics[width=.24\linewidth]{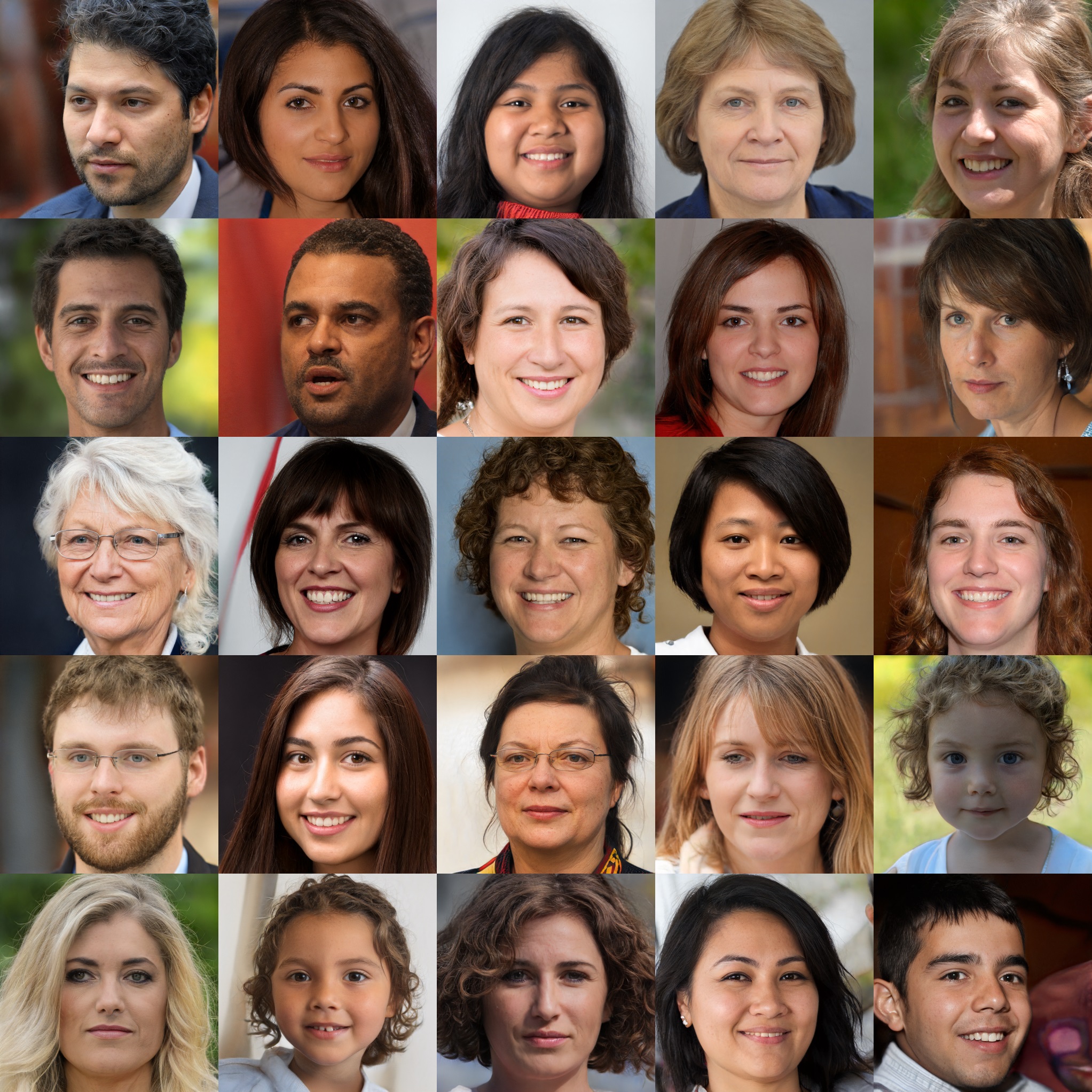}}
        \,
    \subfloat[\centering $k = 5$ ]{\includegraphics[width=.24\linewidth]{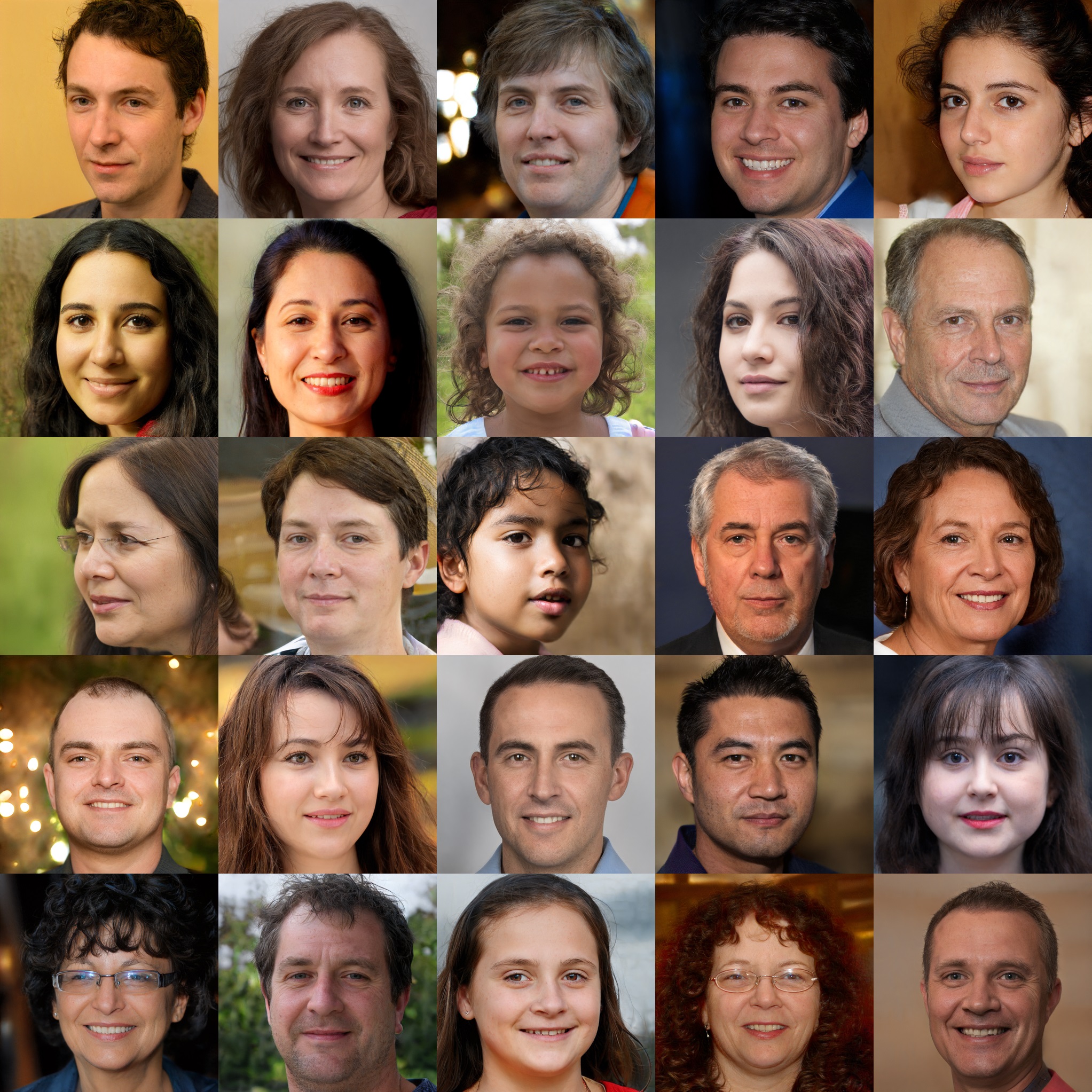}}
        \,
    \subfloat[\centering $k = 7$ ]{\includegraphics[width=.24\linewidth]{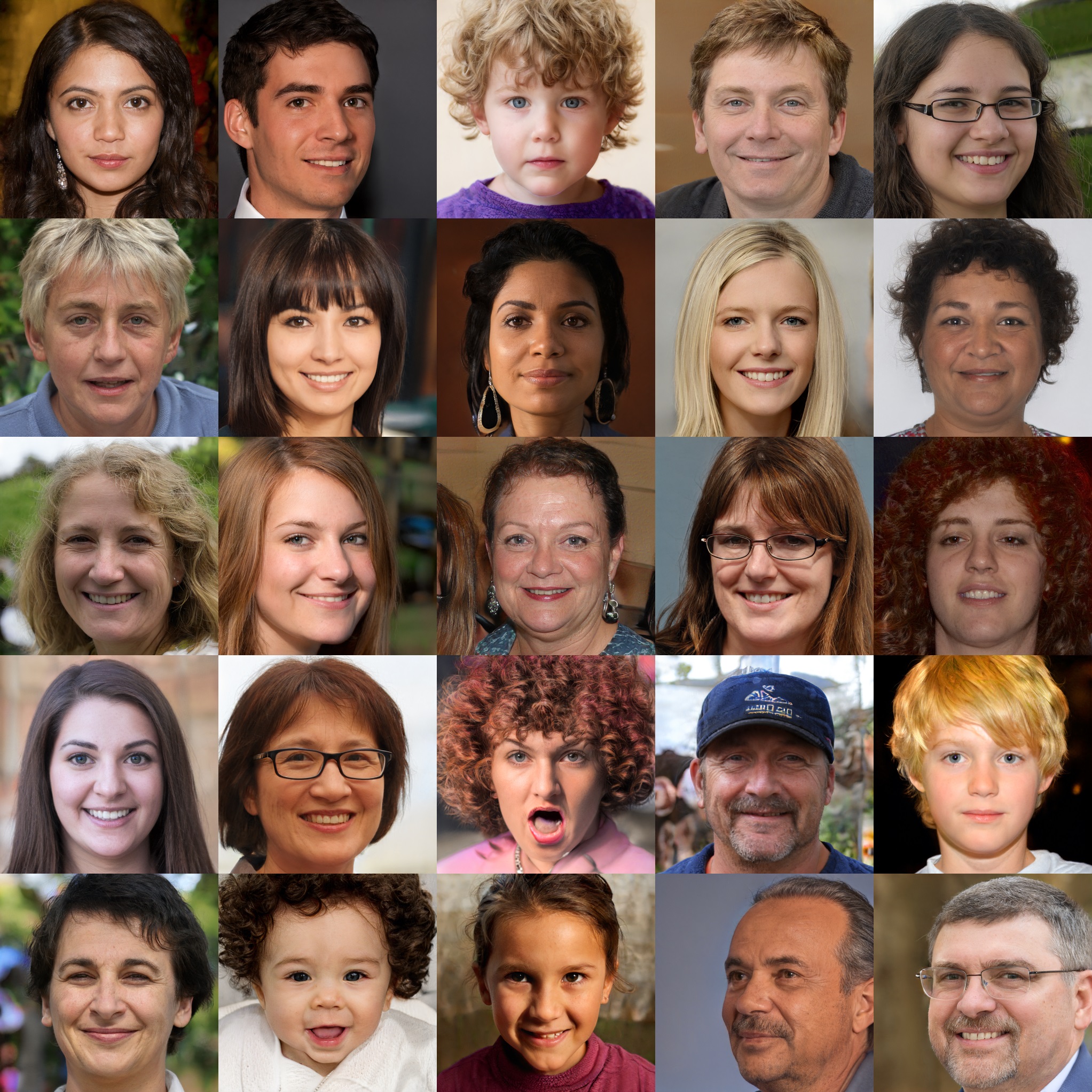}}
        \,
    \subfloat[\centering $k = 10$ ]{\includegraphics[width=.24\linewidth]{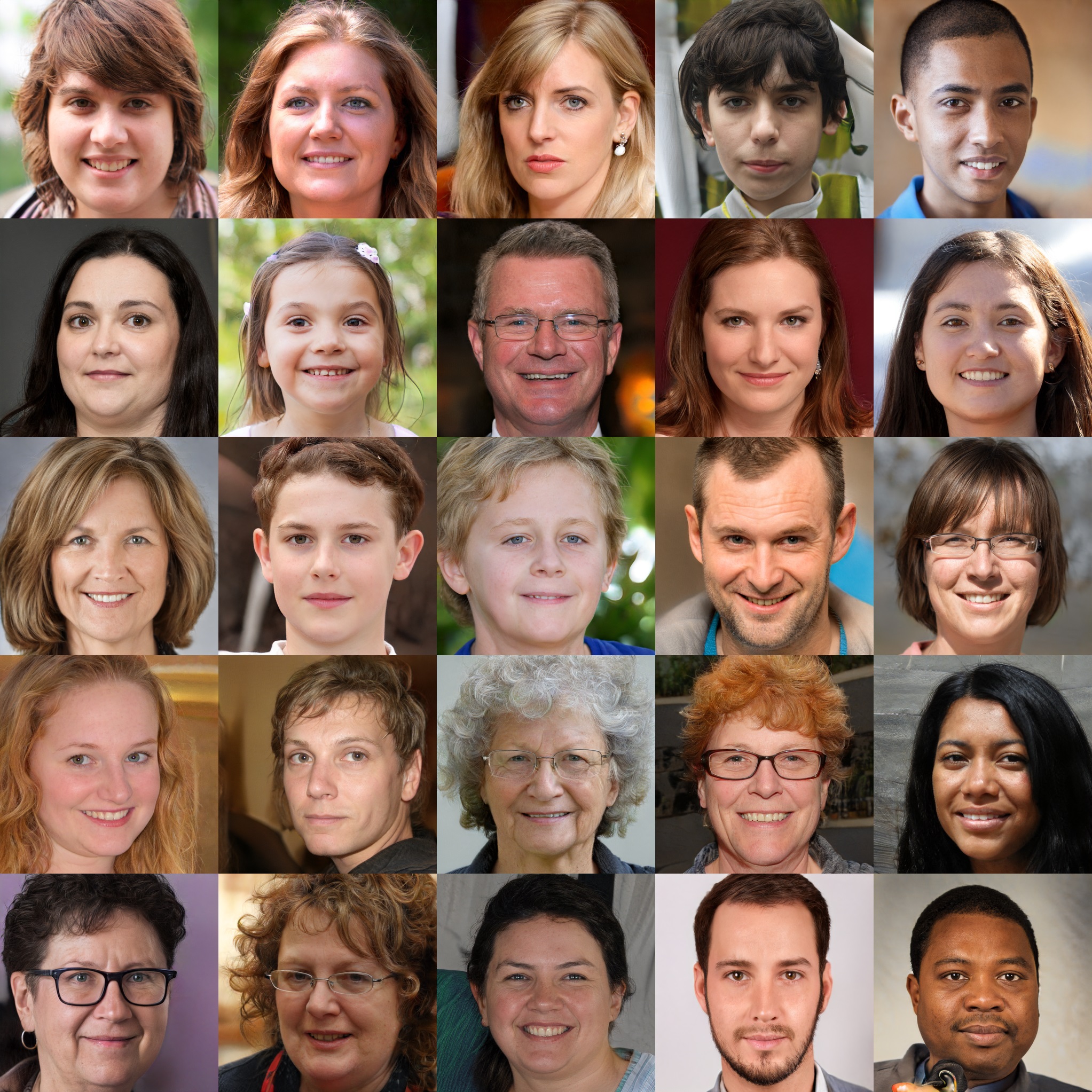}}
    
    \caption{Performance of our method with different choices of hyper-parameter $k=3,5,7,10$. We use $t=50$, $n=10000$.}%
    \label{fig:diffK}%
    \vspace{-4mm}
\end{figure*}

\subsection{Justification of Algorithmic Choices}

\noindent \textbf{Threshold $t$.}
In our method, given a fixed latent sample set $S$, the threshold $t$ determines the trade-off\footnote{Note that this trade-off only applies to a fixed $S$. Our method can generate an infinite number of high-quality samples by simply using multiple latent sets $S_1$, $S_2$, $...$, $S_N$ or a larger $S$.} between image quality and number of output latents: the larger $t$, the higher image quality, but the fewer output images.
However, as Fig.~\ref{fig:diffTreshold} and Table~\ref{table:fid_algChoice} show, we observed that the image quality remains high for various choices of $t$. 
Since the image quality is not sensitive to the choice of $t$ in a relatively large range, we suggest using $t=50$ as the default value for the case when $n=10000$, $k=5$.
Note that we can easily extend our algorithm to output a user-specified number of images (denoted as $n'$) by using a revised scheme: 
if there are enough images in $S$, we first sort all images in the descending order of hub value $m$, and keep the top $n'$ latents; otherwise, we successively draw more latent sets $S_{i}$ and keep all $m>t$ images from them until we get $n'$ images.

\vspace{2mm}
\noindent \textbf{Hyper-parameter $k$.}
We tested the performance of our algorithm with various choices of $k = 1, 3, 5, 7, 10$ in the $k$-NN algorithm. 
Apart from the case when no hub latents can be found ($k = 1$), we show the results of $k = 3, 5, 7, 10$ in Fig.~\ref{fig:diffK} and Table~\ref{table:fid_algChoice}.
It can be observed that the image quality is not sensitive to the choice of $k$.
Nevertheless, we noticed that using a larger $k$ yields more output hub latents for a given latent set $S$ and threshold $t$, but at the cost of slightly longer computation (Appendix~\ref{appendix:running_time}).
To achieve a balance, we suggest using $k=5$ as a default value when $n=10000$, $t=50$.

\vspace{2mm}
\noindent \textbf{Size of Latent Sample Set $n$.}
As Table~\ref{table:fid_algChoice} shows, we also test the performance of the proposed method against various sizes $n=10000, 20000, 30000, 40000$ of latent sample set $S$. Please see Appendix~\ref{appendix:frequency_distribution_additional} for qualitative results.
Similar to above, we observed that (i) although the FID scores get slightly better with increasing $n$, the image quality is not sensitive to the choice of $n$; (ii) using a larger $n$ yields more output hub latents but at the cost of longer computation (Appendix~\ref{appendix:running_time}).
To achieve a balance, we suggest using $n=10000$ as a default value when $k=5$, $t=50$.

\begin{table}[]
\centering
\vspace{-2mm}
\caption{FID scores of StyleGAN2 images synthesized using our method with different choices of $k$, $t$ and $n$, whose default values are $k=5$, $t=50$ and $n=10000$.
We sample 2,000 images to compute the FIDs, whose rationale is discussed in Sec.~\ref{sec:truncation_trick}.}
\begin{tabular}{rr|rr|rr}
\toprule
$k$  &   FID$\downarrow$  & $t$  &  FID$\downarrow$   &  $n$      &  FID$\downarrow$  \\ \midrule
$3$  & 22.793 & $60$ & 20.749 &  $10000$  & 22.782\\
$5$  & 22.782 & $50$ & 22.782 &  $20000$  & 22.021\\
$7$  & 22.720 & $40$ & 24.517 &  $30000$  & 21.679\\
$10$ & 22.560 & $35$ & 25.412 &  $40000$  & 19.124\\ \bottomrule
\end{tabular}
\vspace{-6mm}
\label{table:fid_algChoice}
\end{table}

\begin{figure*}[htp]
\centering
    \subfloat[\centering Distance to the mean of all sampled latents ]{\includegraphics[width=.47\linewidth]{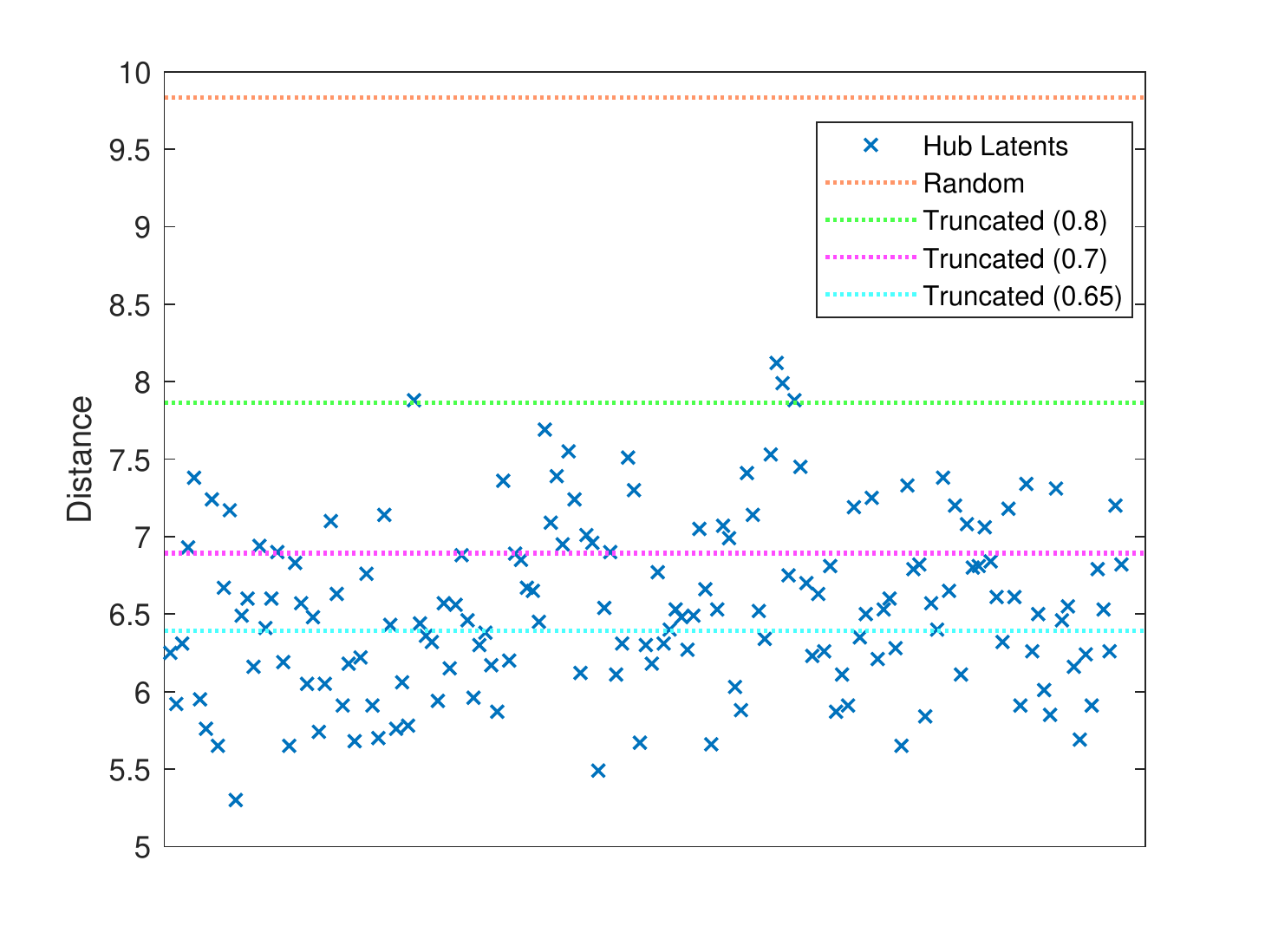}\vspace{-5mm}}%
    \,
    \subfloat[\centering Distance to the mean of hub latents ]{\includegraphics[width=.47\linewidth]{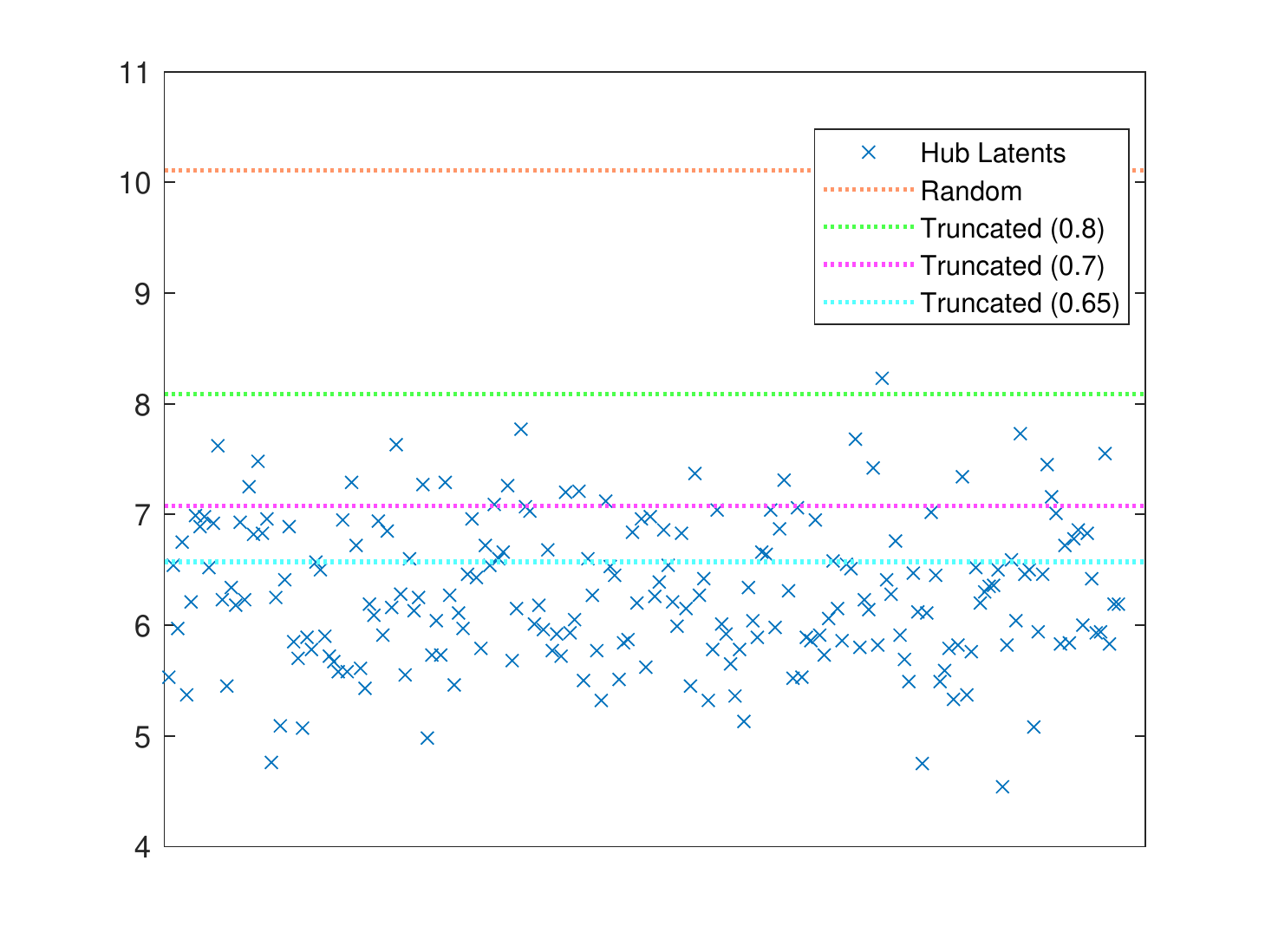}\vspace{-5mm}}
    \caption{The distances of our hub latents to (a) the mean of all sampled latents and (b) the mean of hub latents. Random: the average distance of randomly sampled latents; Truncated ($\psi_0$): the average distance of latents after truncation trick ($\psi=\psi_0$).}%
    \vspace{-2mm}
    \label{fig:hubs_andtrunc3}%
\end{figure*}

\subsection{Relationship with Truncation Trick}
\label{sec:truncation_trick}

\label{sec:comparison_truncation_trick}

The truncation trick~\cite{marchesi2017megapixel,brock2018large,karras2019style} has been widely used in state-of-the-art GANs. Specifically, it truncates randomly sampled latents $\mathbf{w}$ to $\mathbf{w'}=\mathbf{\bar{w}} + \psi (\mathbf{w} - \mathbf{\bar{w}})$ to obtain high-quality latents that yield high-quality synthesized images, where $\mathbf{\bar{w}}$ is the mean of a large number of randomly sampled latents, $\psi$ is a scaling parameter. 
As discussed in Section~\ref{sec:exploitation_hubness}, we argue that it is a naive approximation of our method.

\begin{figure}[htp]
\centering
    \centering
    \subfloat[\centering  Hub latents (distant) ]{\includegraphics[width=.9\linewidth]{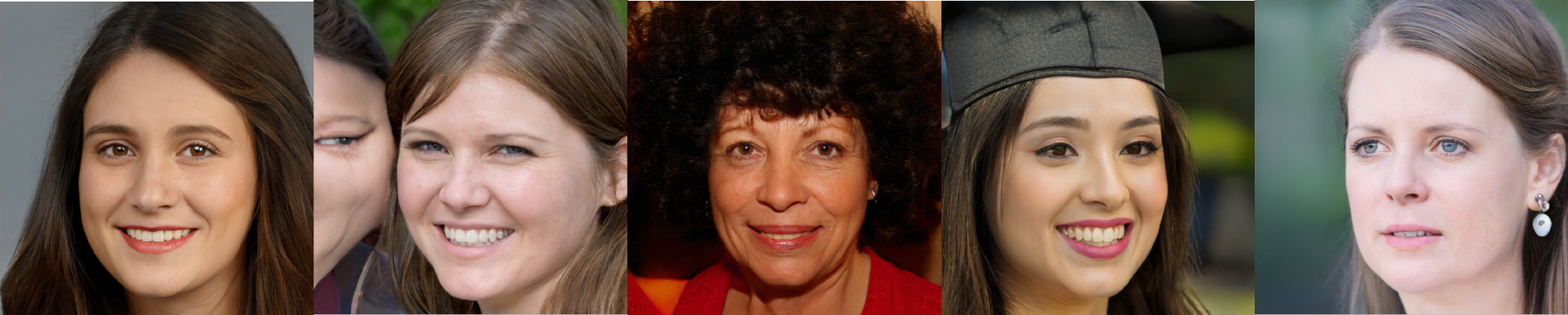}
    }%
    \qquad
    \subfloat[\centering Truncated latents ($\psi = 0.8$)] {\includegraphics[width=.9\linewidth]{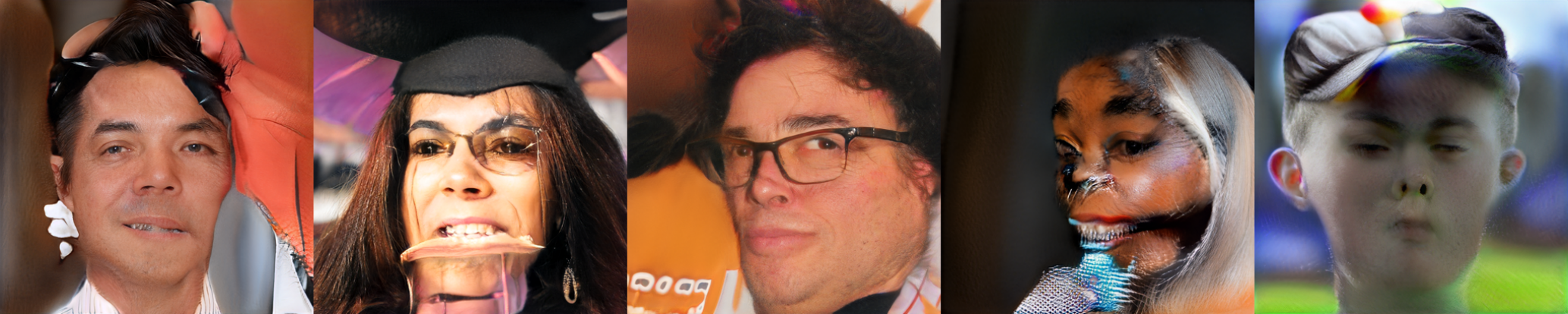}}
    \caption{StyleGAN2 images synthesized from (a) distant hub latents far from their mean; (b) truncated latents ($\psi = 0.8$).}%
    \vspace{-4mm}
    \label{fig:comp_dist}%
\end{figure}

\noindent \textbf{Distance to the Means of Hub and All Latents.}
To justify our claim, we first investigate the distances of our {\it hub latents} to their mean and their distances to the mean of all sampled latents.
As Fig.~\ref{fig:hubs_andtrunc3} shows, it can be observed that: i) Our hub latents are closer to both the hub mean and the all latent mean than randomly sampled latents, which justifies the ``central clustering effect'' of our hub latents~\cite{radovanovic2010hubs}. ii) Surprisingly, the distances of most hub latents are around $6.0$ to $7.0$ for both cases, which is roughly the same as the distances of randomly sampled latents truncated with a parameter $\psi =0.7$, {\it i.e.} the StyleGAN-recommended~\cite{karras2019style} parameter value for the truncation trick. However, StyleGAN obtained the value $\psi =0.7$ empirically via try-and-error while we obtain it as a byproduct of our method, which justifies the superiority and fundamentality of our approach.
iii) A small portion of our hub latents are of larger distances ({\it e.g.} around $7.5$ and $8.0$) to the means, which will be overlooked by the truncation trick with $\psi =0.7$. In addition, applying the truncation trick with $\psi =0.8$ are prone to get low-quality latents that yield low-quality images while our ``distant'' hub latents are still of high quality (Fig.~\ref{fig:comp_dist}). This further justifies the superiority of our method against the truncation trick.

\begin{table}[t]
\centering
\vspace{-4mm}
\caption{Comparison of FID scores of StyleGAN2 synthesized images using our method and the truncation trick. FFHQ-1 and FFHQ-2: real images sampled from the FFHQ dataset~\cite{karras2019style}; Hubs ($50$): our method with $t=50$; Truncated ($0.7$): truncation trick with $\psi=0.7$; Random: random sampling. We sample $2000$ latents/images for all methods compared. The FID scores between i) FFHQ-1 and FFHQ-2; and ii) Random and FFHQ-1,FFHQ-2 are used as baselines. Dist2Mean: distances of sampled latents to the all latent mean.}
\begin{tabular}{l r r r}
\toprule
\multirow{2}{*}{Methods}  & \multicolumn{2}{c}{FID$\downarrow$} & \multirow{2}{*}{Dist2Mean}\\
                 & \multicolumn{1}{r}{FFHQ-1} & FFHQ-2  & \\ 
                 \hline
FFHQ-2 & \multicolumn{1}{r}{16.505} &    ----         & ----            \\ 
Hubs ($50$) & \multicolumn{1}{r}{21.955} & 23.609 & 6.247   \\
Truncated ($0.7$) & \multicolumn{1}{r}{25.097} & 25.127 & 6.893    \\ Random & \multicolumn{1}{r}{35.455} & 35.598 & 9.847   \\
\bottomrule
\end{tabular}
\vspace{-4mm}
\label{tab:FIDandDist}
\end{table}

\begin{figure*}[htp]
\centering
    \begin{subfigure}[h]{0.32\textwidth}
      \centering
        \includegraphics[width=0.99\linewidth]{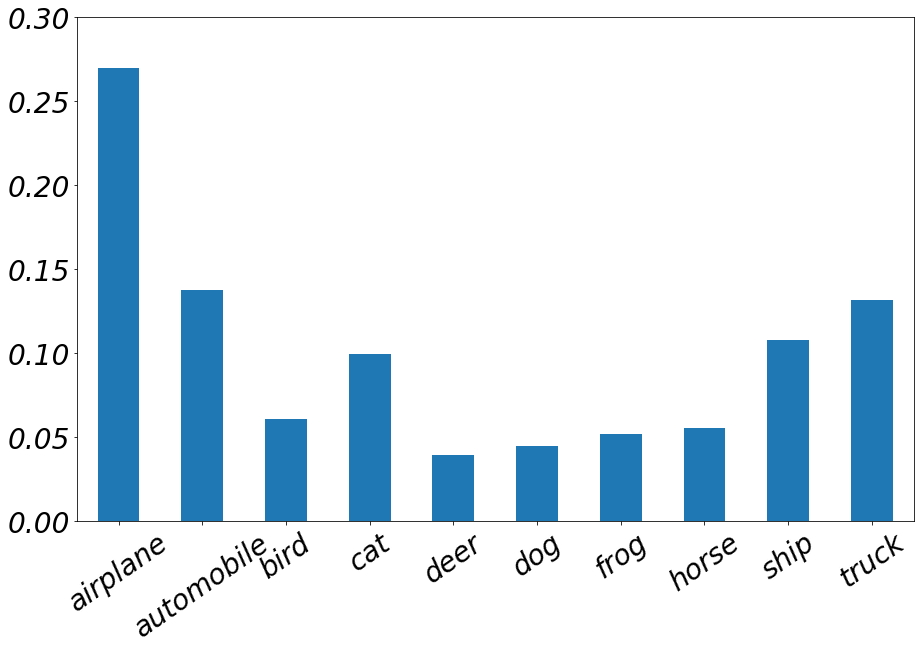}%
      \caption{Hubs ($50$), $\mathrm{WD}=0.0094$}
    \end{subfigure}
    \begin{subfigure}[h]{0.32\textwidth}
        \centering
        \includegraphics[width=0.99\linewidth]{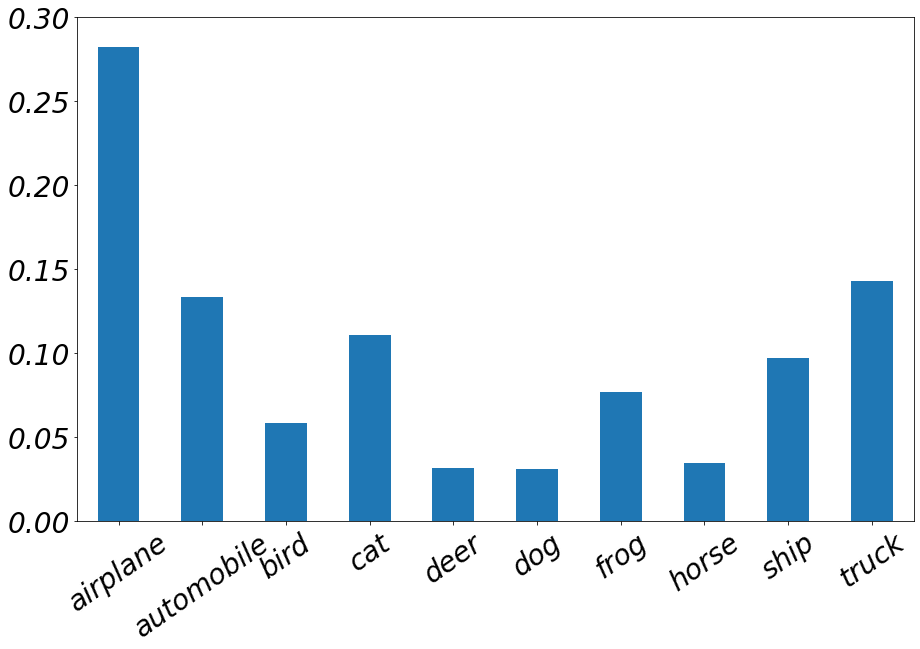}%
      \caption{Truncated ($0.7$), $\mathrm{WD}=0.0102$}
    \end{subfigure}
    \rulesep
    \begin{subfigure}[h]{0.32\textwidth}
       \includegraphics[width=.99\linewidth]{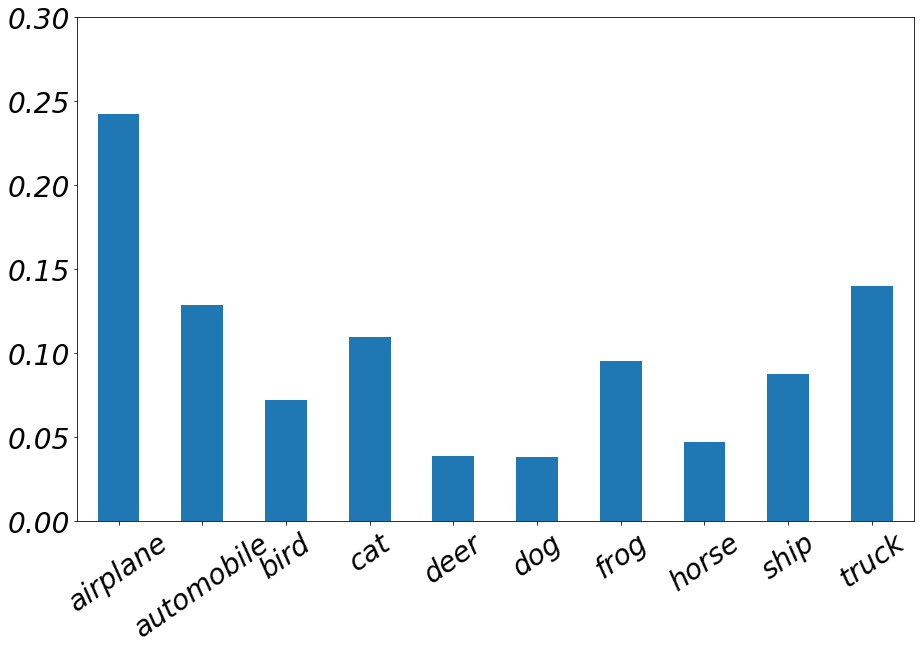}%
        \caption{Random}
    \end{subfigure}
    \caption{The class distributions of the StyleGAN2 model pretrained on the CIFAR10 dataset with (a) our hubness-based sampling ($t=50$), (b) the truncation trick ($\psi=0.7$), and (c) the random sampling methods. WD: Wasserstein distance.}
   \label{fig:multi-distribution}
   \vspace{-4mm}
 \end{figure*}

\vspace{2mm}
\noindent \textbf{FID Scores.}
As Table~\ref{tab:FIDandDist} shows, we also justify the superiority of our method by comparing the FID scores~\cite{heusel2017gans} of images generated by StyleGAN2 using both the truncation trick, $\psi = 0.7$~\cite{karras2019style} and our method.
Specifically, we compute the FID scores between images generated by i) real images sampled from the FFHQ datasets, {\it i.e.} FFHQ-1 and FFHQ-2 in Table~\ref{tab:FIDandDist}; ii) our hub latents and FFHQ-1, FFHQ-2; iii) truncated latents ($\psi=0.7$) and FFHQ-1, FFHQ-2; iv) randomly sampled latents and FFHQ-1, FFHQ-2.
It can be observed that i) both our method and the truncation trick outperform random sampling; ii) our method achieves better FID scores than the truncation trick.
Note that we intentionally used a small number of images ({\it i.e.} 2,000) to compute FID to avoid covering the entire distribution and thus suffer less from the restriction of latent spaces.
In comparison with the results in~\cite{jung2021internalized} and the bias-free FID$_{\infty}$~\cite{chong2020effectively} computed with 10K images (Table~\ref{table:new_fid}), our FID scores of ``Truncated (0.7)'' images are better than ``Random'', which is consistent with human perception.
Note that our method outperforms Truncated (0.7) in both cases.
Examples of StyleGAN2 synthesized images after the truncation trick ($\psi=0.7$) are shown in Appendix~\ref{appendix:truncation_trick_example}.
Nevertheless, even using a small number of images, FID may still not be a good evaluation metric for our task. Therefore, we resort to the precision and recall metrics~\cite{kynkaanniemi2019improved} that make more sense.

\begin{table}[h]
\centering
\vspace{-4mm}
\caption{FID$_{\infty}$ scores~\cite{chong2020effectively} computed with 10K images, which are ineffective as they capture the entire distribution and thus suffer from the restriction of latent spaces. \textcolor{red}{Red}: random sampling has the best score, which contradicts human perception as the images sampled with it are of the lowest quality.}
\begin{tabular}{l|r r r}
\toprule
Method  & Hubs ($50$) & Truncated ($0.7$) & Random \\ 
\midrule
FID$_{\infty}\downarrow$ & {\bf 15.398} & 15.761 & \textcolor{red}{2.923}\\
\bottomrule
\end{tabular}
\vspace{-2mm}
\label{table:new_fid}
\end{table}

\vspace{2mm}
\noindent \textbf{Precision and Recall~\cite{kynkaanniemi2019improved}.}
As Table~\ref{table:Pre_Recall} shows, our method achieves a high precision comparable to Truncated ($0.3$) which sacrifices the synthesis diversity ({\it i.e.} low recall) while retaining a very high recall comparable to Random which includes many low-quality results ({\it i.e.} low precision). 
This further justifies the superiority of our method.

\begin{table}[h]
\centering
\vspace{-4mm}
\caption{Comparison of precision and recall ~\cite{kynkaanniemi2019improved} of StyleGAN2 synthesized images using our method and the truncation trick.}
\begin{tabular}{l|r|r|r}
\toprule
Method  &   Precision$\uparrow$  & Recall$\uparrow$ \\ \midrule 
Hubs ($50$)         & \underline{0.890} & \underline{0.324} \\
Truncated ($0.3$)  & {\bf 0.892} & 0.015 \\
Truncated ($0.7$)  & 0.811 & 0.223 \\
Random  & 0.720 & {\bf 0.393} \\
\bottomrule
\end{tabular}
\vspace{-4mm}
\label{table:Pre_Recall}
\end{table}

\subsection{Impact on Class Balance}

We further investigate how our method affects the class balance of unconditional GANs pre-trained on multi-class datasets.
As Fig.~\ref{fig:multi-distribution} shows, we evaluate the class balance of a StyleGAN2 model pretrained on the CIFAR10 dataset with i) random sampling\footnote{\url{https://github.com/POSTECH-CVLab/PyTorch-StudioGAN}} ({\it i.e.} Random), ii) truncation trick ($\psi=0.7$) and iii) our hubness-based sampling method. Specifically, we sample 50,000 images each and use a pretrained CIFAR10 classifier\footnote{\url{https://github.com/open-mmlab/mmclassification}, ResNet50} to estimate their class distributions.
Note that although a ``larger'' difference can be observed visually, our method actually preserves the class balance better as it has a smaller Wasserstein distance to the distribution of Random than the truncation trick.
In addition, as Table~\ref{table:multi} shows, our method achieves a better Inception Score~\cite{salimans2016improved} that favours balanced and high-confidence classifications, which further justifies the superiority of our method in preserving class balance.

\begin{table}[h]
\centering
\vspace{-4mm}
\caption{Evaluation of class balance with Inception Scores (IS) ~\cite{salimans2016improved} of StyleGAN2 pretrained on the CIFAR10 dataset using our hubness-based sampling ($t=50$), the truncation trick ($\psi=0.7$), and the random sampling methods.}

\begin{tabular}{l|r r r}
\toprule
Method  &   Hubs ($50$)  & Truncated ($0.7$) & Random\\ 
\midrule
IS         & \textbf{6.212} &  6.059 & 7.080\\

\bottomrule
\end{tabular}
\vspace{-4mm}
\label{table:multi}
\end{table}

\section{Conclusions}

In this paper, we address the quality variance of GAN synthesized images by investigating the sampling of GAN latents.
Specifically, we first show that GAN latents are not uniformly distributed in the latent space due to the {\it hubness} phenomenon of data distributions in high dimensional space. In addition, there exist {\it hub} latents that are much more likely to be nearest neighbors of others and contribute more to the synthesis of high-quality images.
Then, we formulate the above as the {\it hubness priors} and propose a novel GAN latent sampling algorithm, which allows for efficient and high-quality image synthesis for GANs.
Furthermore, we show that the well-known truncation trick is a naive approximation of our method that utilizes the ``central clustering effect'' of {\it hub} latents, which not only uncovers the rationale of the truncation trick, but also indicates that our method is superior and more fundamental.

\section*{Acknowledgements}

We appreciate the reviewers’ constructive comments in improving the paper.
This research was partially funded by the UK Engineering and Physical Sciences Research Council (EPSRC) through the Doctoral Training Partnerships (DTP) with No. EP/T517951/1 (2599521).

\bibliography{example_paper}
\bibliographystyle{icml2022}

\appendix

\section{Additional Experimental Results}
\label{appendix:frequency_distribution_additional}

\begin{figure}[htp]
\centering
    \centering
    \subfloat[\centering StyleGAN-Horse-HQ]{\includegraphics[width=.45\linewidth]{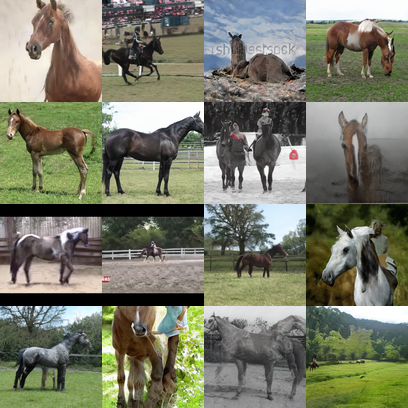}}%
    \qquad
    \subfloat[\centering StyleGAN-Horse-LQ
    ]{\includegraphics[width=.45\linewidth]{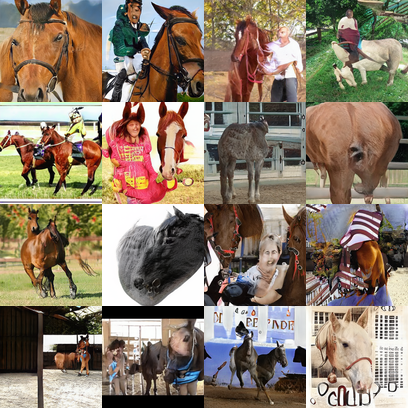}}%
    
    \caption{Performance of our method on StyleGAN2 pretrained on the horse domain. It can be observed that our method works well.
    (a) and (b) are images synthesized from high-quality (HQ) and low-quality (LQ) latents obtained by Algorithm~\ref{alg:hq_sampling} and Algorithm~\ref{alg:lq_sampling} (Appendix~\ref{appendix:pseudocode_lq_sampling}), respectively. We use $n=10000$, $k=5$, $t=50$ and $t_{lq}=1$.}%
    \label{fig:otherdataset_appendix}%
    \vspace{-4mm}
\end{figure}

\vspace{2mm}
\noindent \textbf{Different Image Domains.} 
As Fig.~\ref{fig:otherdataset_appendix} shows, as complement to Fig.~\ref{fig:otherdataset} in the main paper, to justify that our method works across different image domains, we show that our method also works on a StyleGAN2 model pretrained on the horse domain.

\begin{figure*}[htp]
 \centering
 \begin{subfigure}[h]{\textwidth}
    \centering
    \includegraphics[width=.32\linewidth]{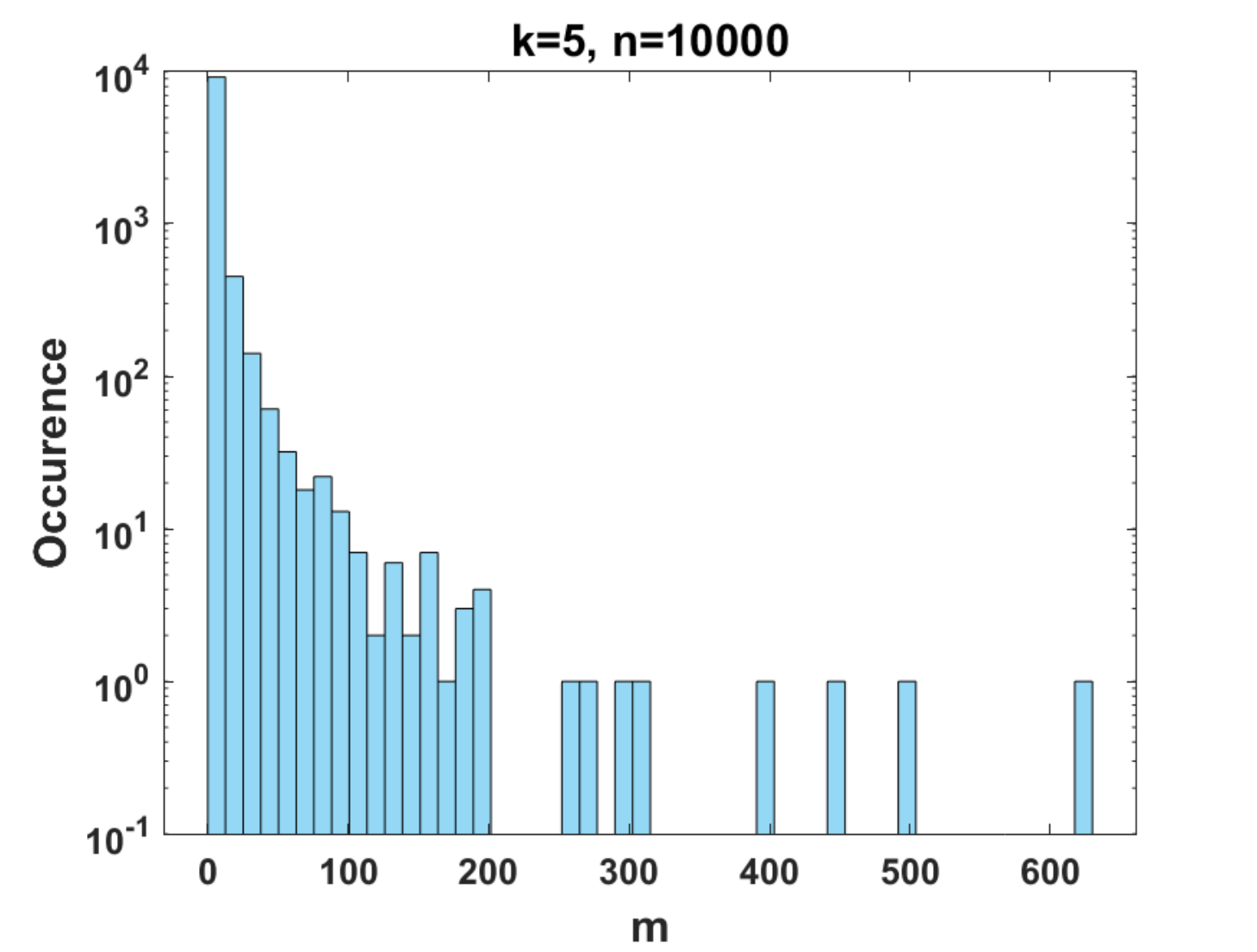}%
    \includegraphics[width=.32\linewidth]{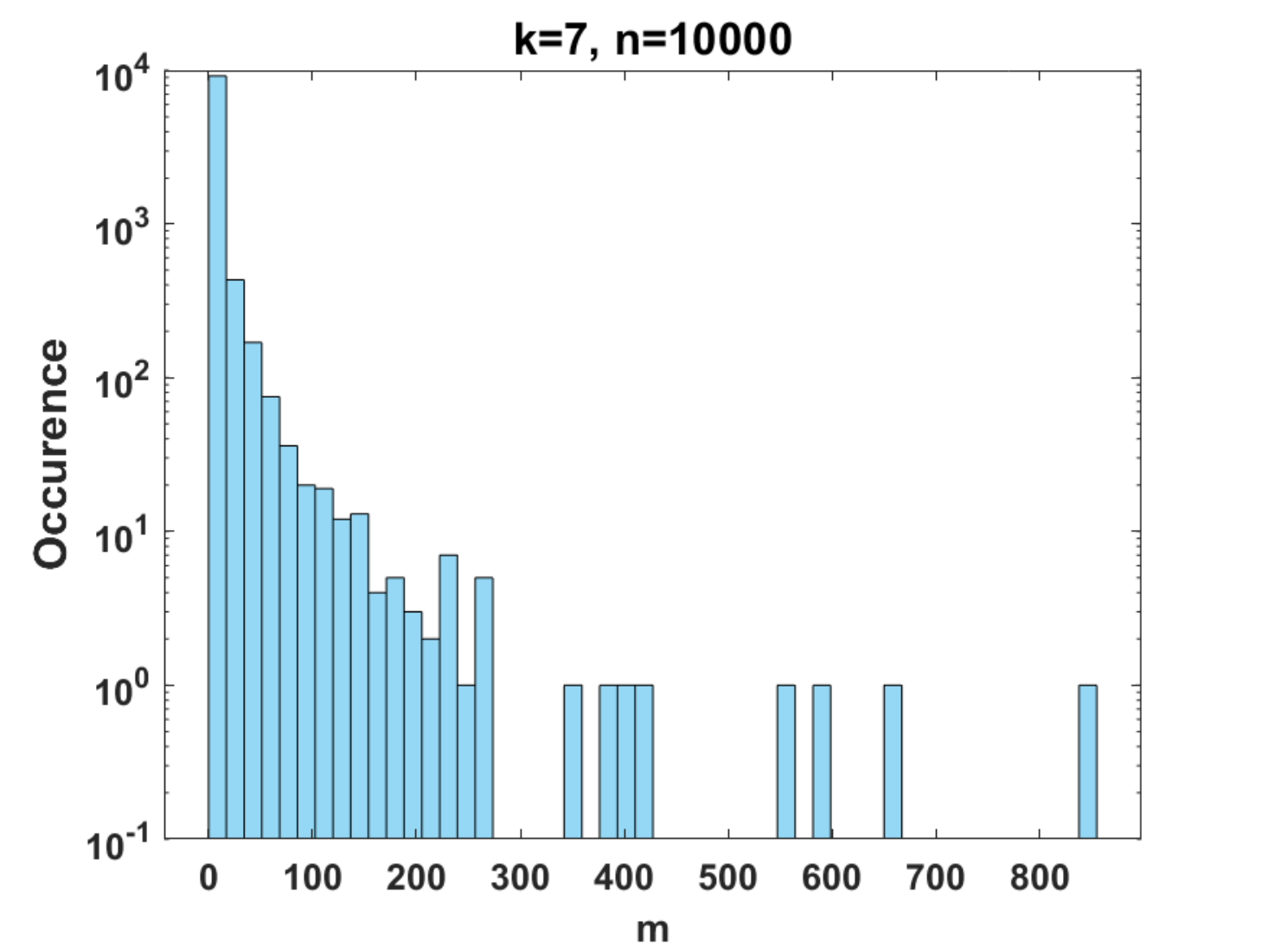}%
   \includegraphics[width=.32\linewidth]{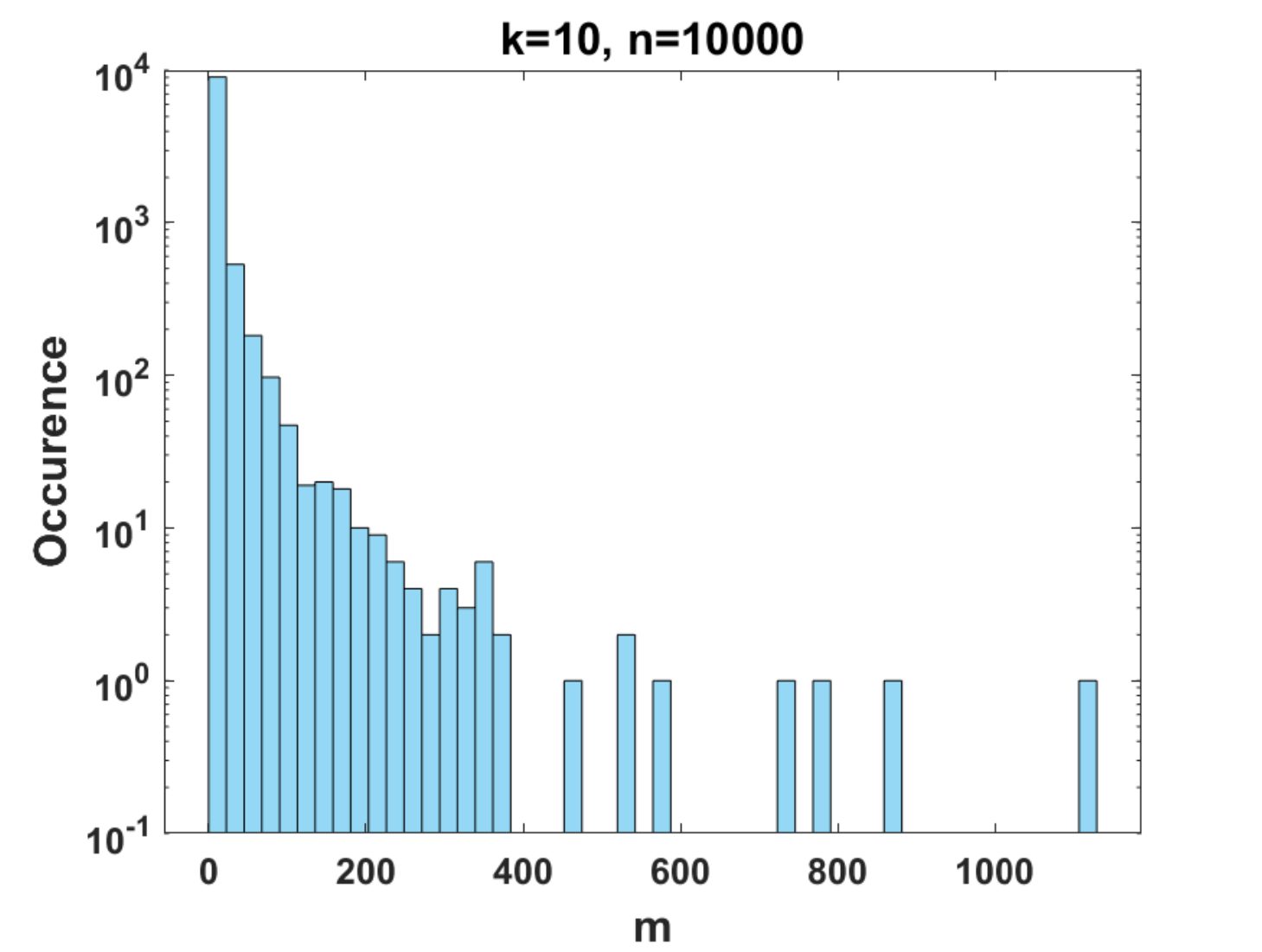}%
   \caption{StyleGAN series~\cite{karras2019style,karras2020analyzing,karras2021alias}, $Z$-space ($512$ dimensions)}
 \end{subfigure}
 \begin{subfigure}[h]{\textwidth}
    \centering
    \includegraphics[width=.32\linewidth]{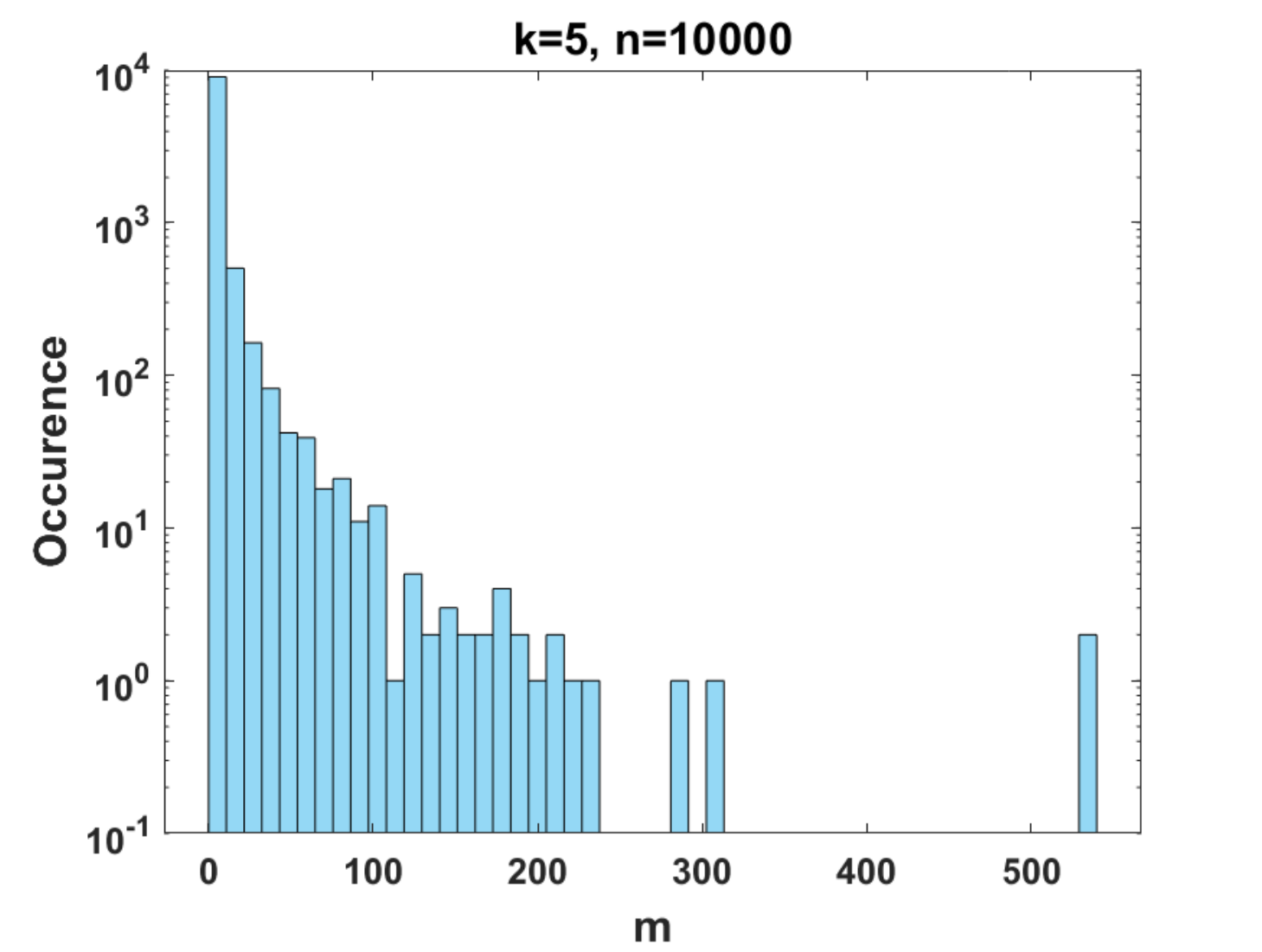}%
    \includegraphics[width=.32\linewidth]{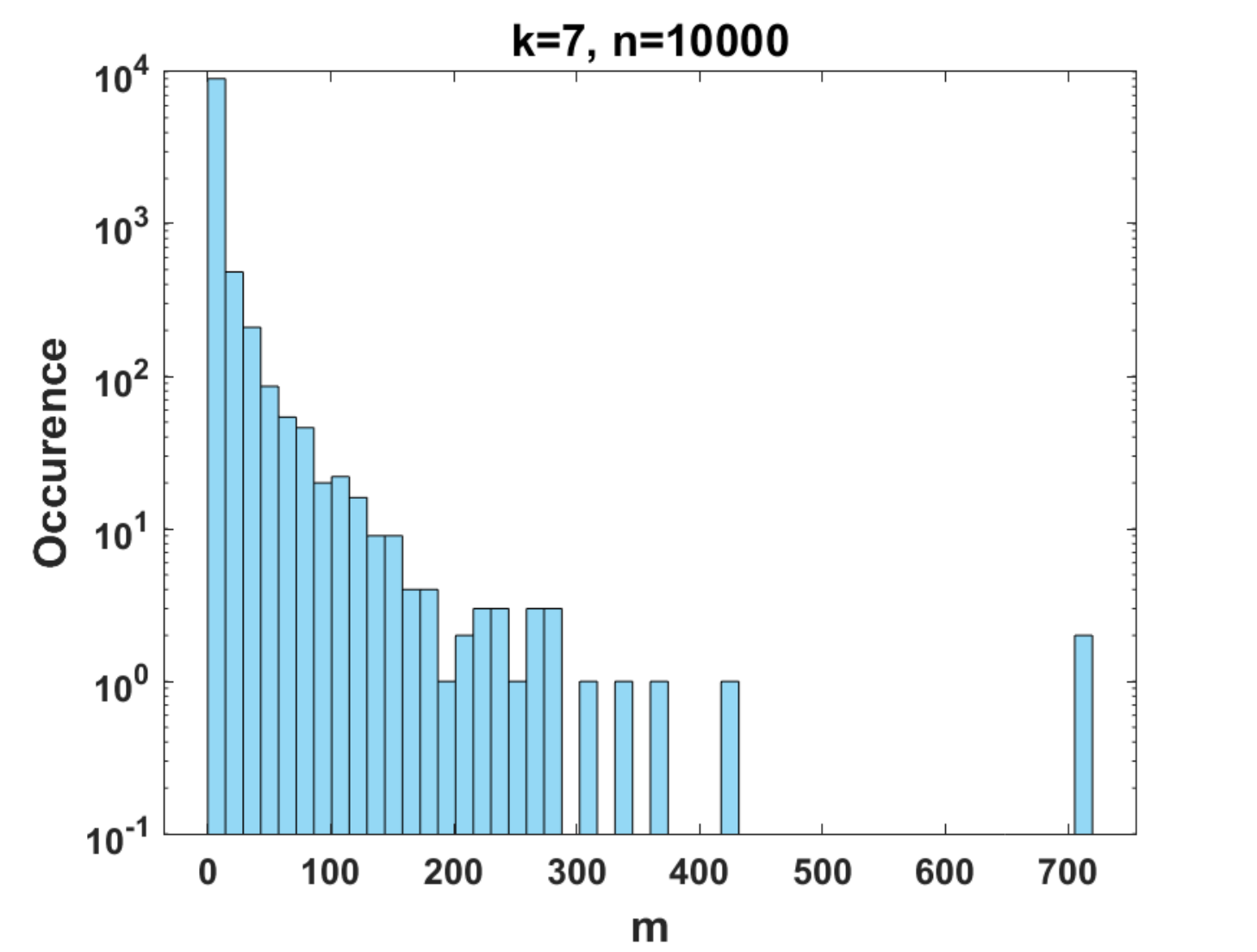}%
   \includegraphics[width=.32\linewidth]{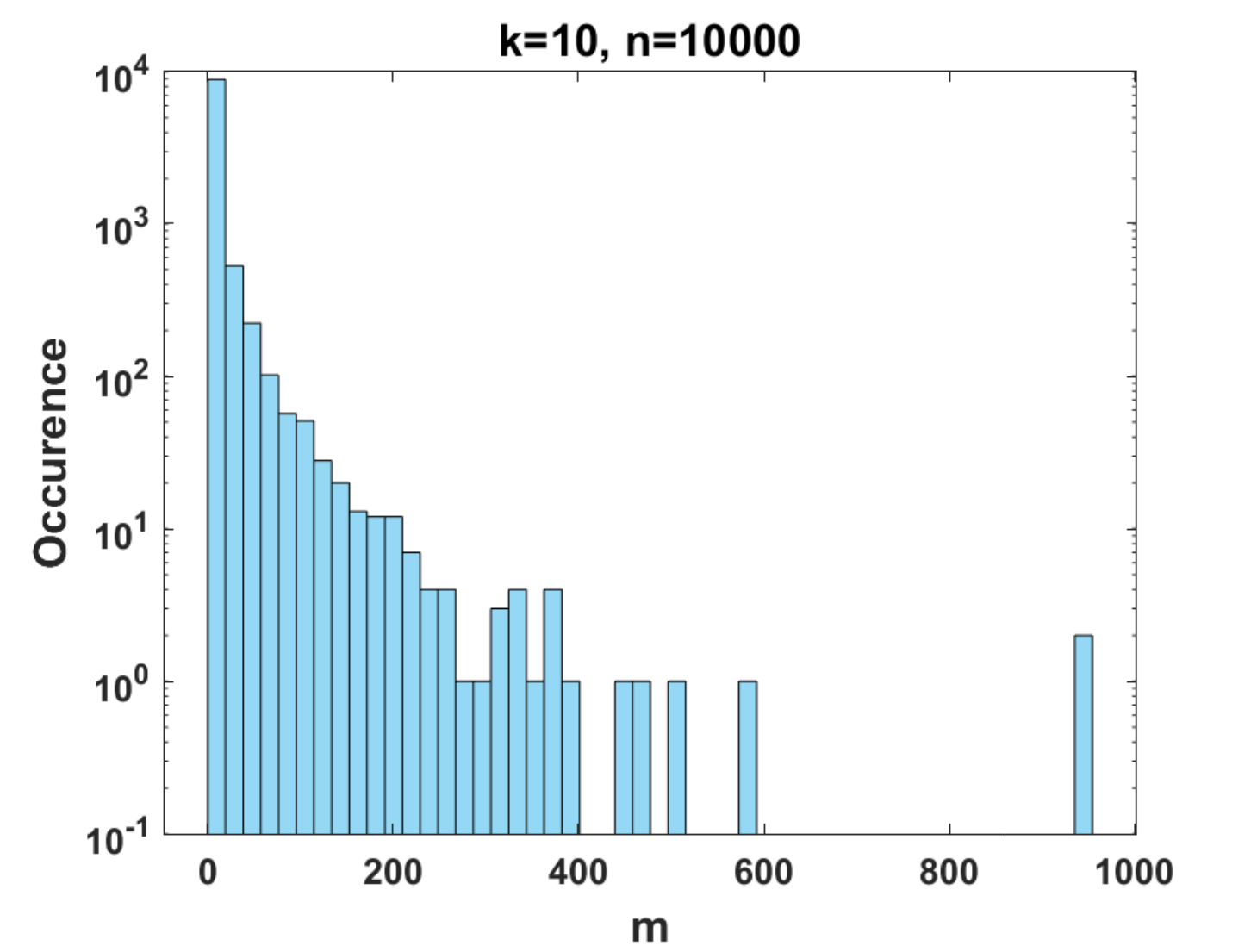}%
   \caption{ProGAN~\cite{karras2018progressive}, $Z$-space$^*$ ($512$ dimensions)}
 \end{subfigure}
    \caption{Distributions of $m$-hub latents for state-of-the-art GANs, $k=5,7,10$ (the $k$-NN algorithm) and $n=10000$ (size of latent sample set $S$). $^*$: Although both are $512$-dimensional, the ProGAN~\cite{karras2018progressive} latents are sampled directly from $\mathcal{N}(0,I)$ while the StyleGAN latents further normalized the sampled latents to be of the same norm~\cite{karras2019style}.
    All distributions are highly tailed to the right, which demonstrates the existence of hubness phenomenon~\cite{radovanovic2010hubs} in GAN latent spaces.}
    \label{fig:exploration_hubness_cont}
\end{figure*}

\noindent \textbf{Distributions of Hub Latents.} 
As a complement to Fig.~\ref{fig:exploration_hubness}, Fig.~\ref{fig:exploration_hubness_cont} shows additional results on the distributions of $m$-hub latents in the latent spaces of state-of-the-art GANs.

\begin{figure*}[htp]
\centering
    \centering
    \subfloat[\centering $n=20000$]{\includegraphics[width=.32\linewidth]{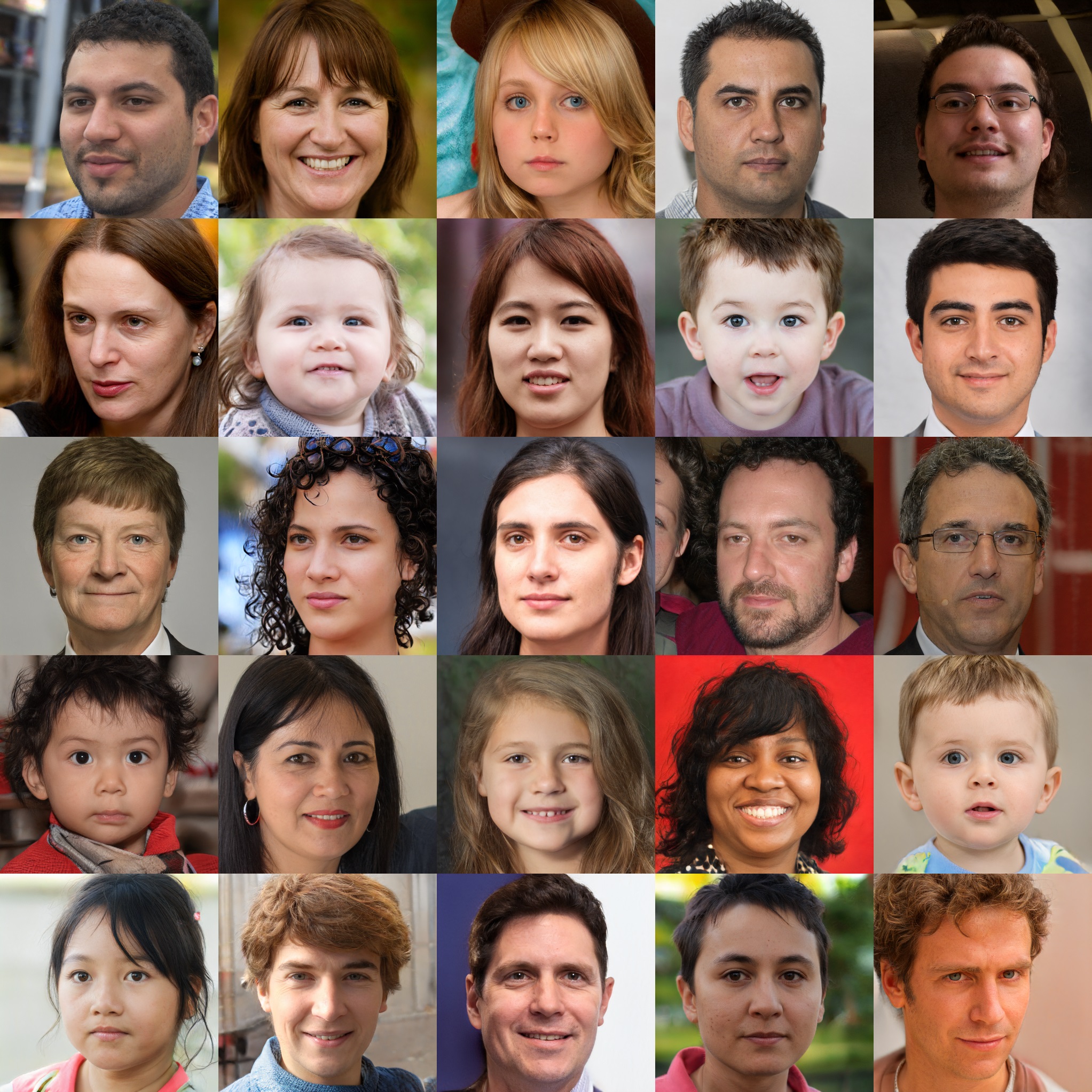}}
    \,
    \subfloat[\centering  $n=30000$]{\includegraphics[width=.32\linewidth]{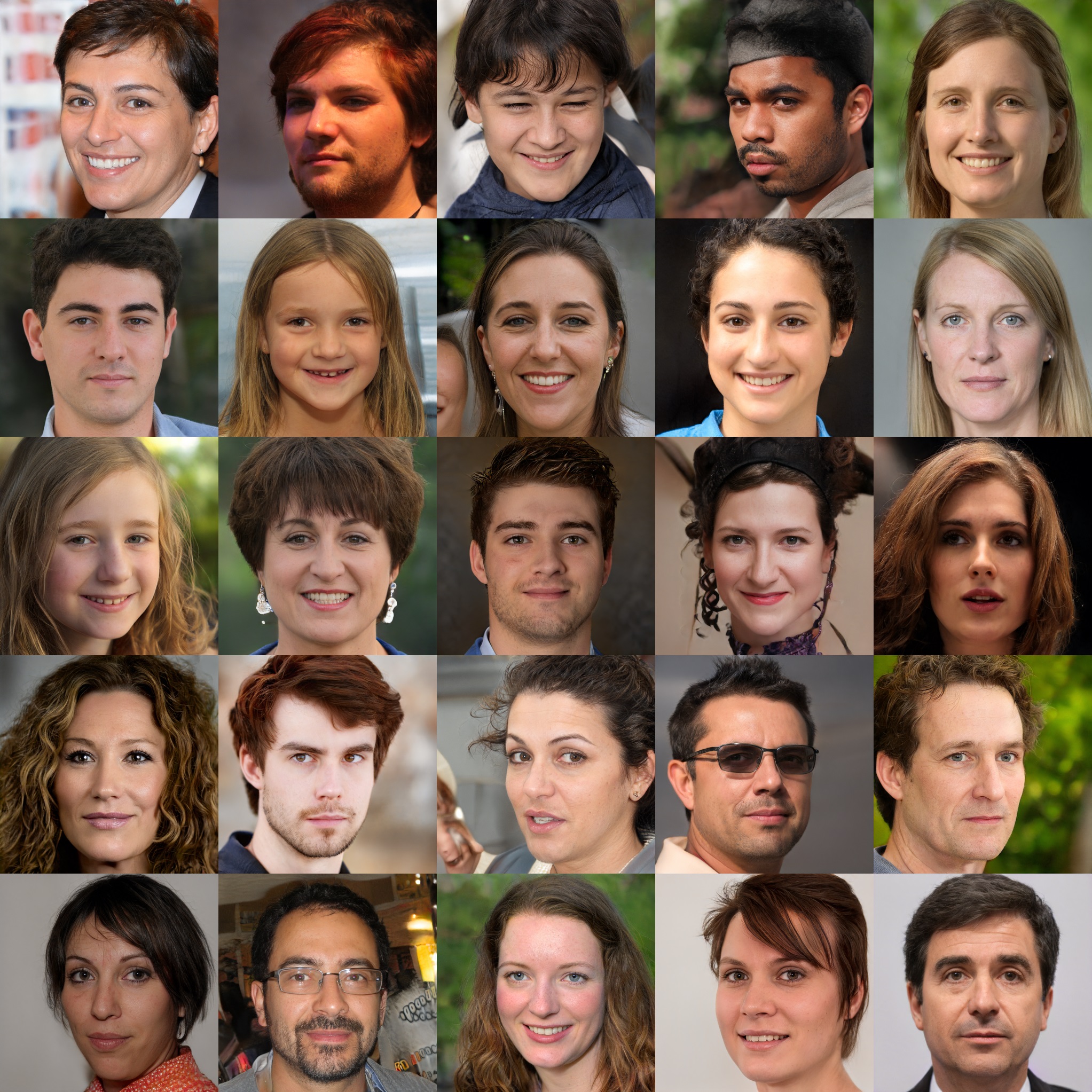}}%
    \,
    \subfloat[\centering  $n=40000$]{\includegraphics[width=.32\linewidth]{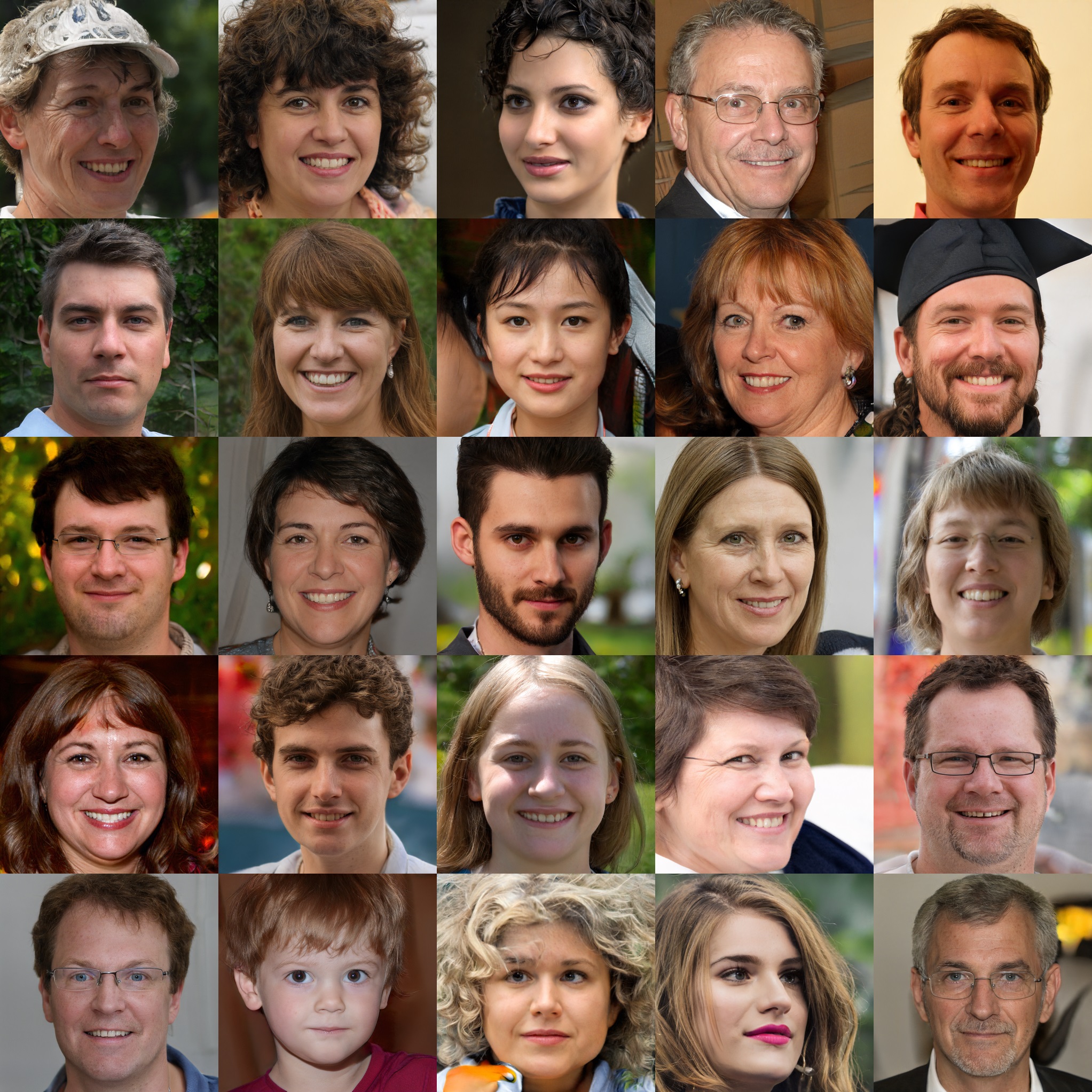}}%
    \caption{Performance of our method with different sizes $n=20000, 30000, 40000$ of sample set $S$. We use $k=5$, $t=50$.}%
    \label{fig:diffsampleS}%
\end{figure*}

\vspace{2mm}
\noindent \textbf{Choice of size of Latent Sample Set $n$.}
Fig.~\ref{fig:diffsampleS} shows the qualitative results of our method with $n=20000$, $30000$, $40000$ (see Fig.~\ref{fig:effectiveness} for results when $n=10000$).
It can be observed that the visual quality remains similar across different $n$, which indicates that the image quality is not sensitive to the choice of $n$.

\vspace{2mm}
\noindent \textbf{Quantitative results with BigGAN~\cite{brock2018large}.}
As Table~\ref{table:biggan} shows, our method outperforms Truncated ($0.7$) with the BigGAN architecture pretrained on the 1000-class ImageNet ILSVRC 2012 dataset on precision and recall~\cite{kynkaanniemi2019improved}, which further justifies the superiority of our method.

 \begin{table}[h]
\centering
\vspace{-6mm}
\caption{Quantitative results with BigGAN (ImageNet).}
\begin{tabular}{l|r|r}
\toprule
Method  &  Precision$\uparrow$ & Recall$\uparrow$ \\ \midrule
Hubs ($50$)  & {\bf 0.147} & {\bf 0.311} \\
Truncated ($0.7$)  &  0.131 & 0.264 \\
\bottomrule
\end{tabular}
\vspace{-4mm}
\label{table:biggan}
\end{table}

\section{Pseudocode of Low-quality GAN Latent Sampling}
\label{appendix:pseudocode_lq_sampling}

The pseudocode of our low-quality GAN latent sampling algorithm is shown in Algorithm~\ref{alg:lq_sampling}. It is a simple inverse of Algorithm~\ref{alg:hq_sampling}, using a different thresholding scheme.

\begin{algorithm}[t]
\textbf{Input:} a set of GAN latents $S=\{z_1, z_2, ..., z_n\}$ sampled from a standard normal distribution $\mathcal{N}(0,I)$, a hyper-parameter $k$, a threshold $t_{lq}$\\
\textbf{Output: $S_{lq}$} 
\begin{algorithmic}
\caption{Low-quality GAN Latent Sampling with Hubness Priors}
\label{alg:lq_sampling}

\STATE \# Step 1
\STATE $m_{1,2,...,n} \gets 0$

\FOR{$i \gets 1 $ to $n$}
    \STATE $\{\mathrm{idx}_1, \mathrm{idx}_2, ...\mathrm{idx}_k\} \gets$ $k$-NN($z_i$)
    \FOR{$j \gets 1 $ to $k$}
        \STATE $m_{\mathrm{idx}_j} \gets m_{\mathrm{idx}_j} + 1$
    \ENDFOR
\ENDFOR
\STATE \# Step 2
\STATE $S_{lq} \gets \emptyset$
\FOR{$i \gets 1 $ to $n$}
    \IF{$m_i \leq t_{lq}$}
        \STATE  $S_{lq} \gets S_{lq} \cup z_{i}$
    \ENDIF
\ENDFOR
\end{algorithmic}
\end{algorithm}

\section{Hubness Spectrum}
\label{appendix:hubs_spectrum}

\begin{figure*}[htp]
    \centering
    \includegraphics[width=.99\textwidth]{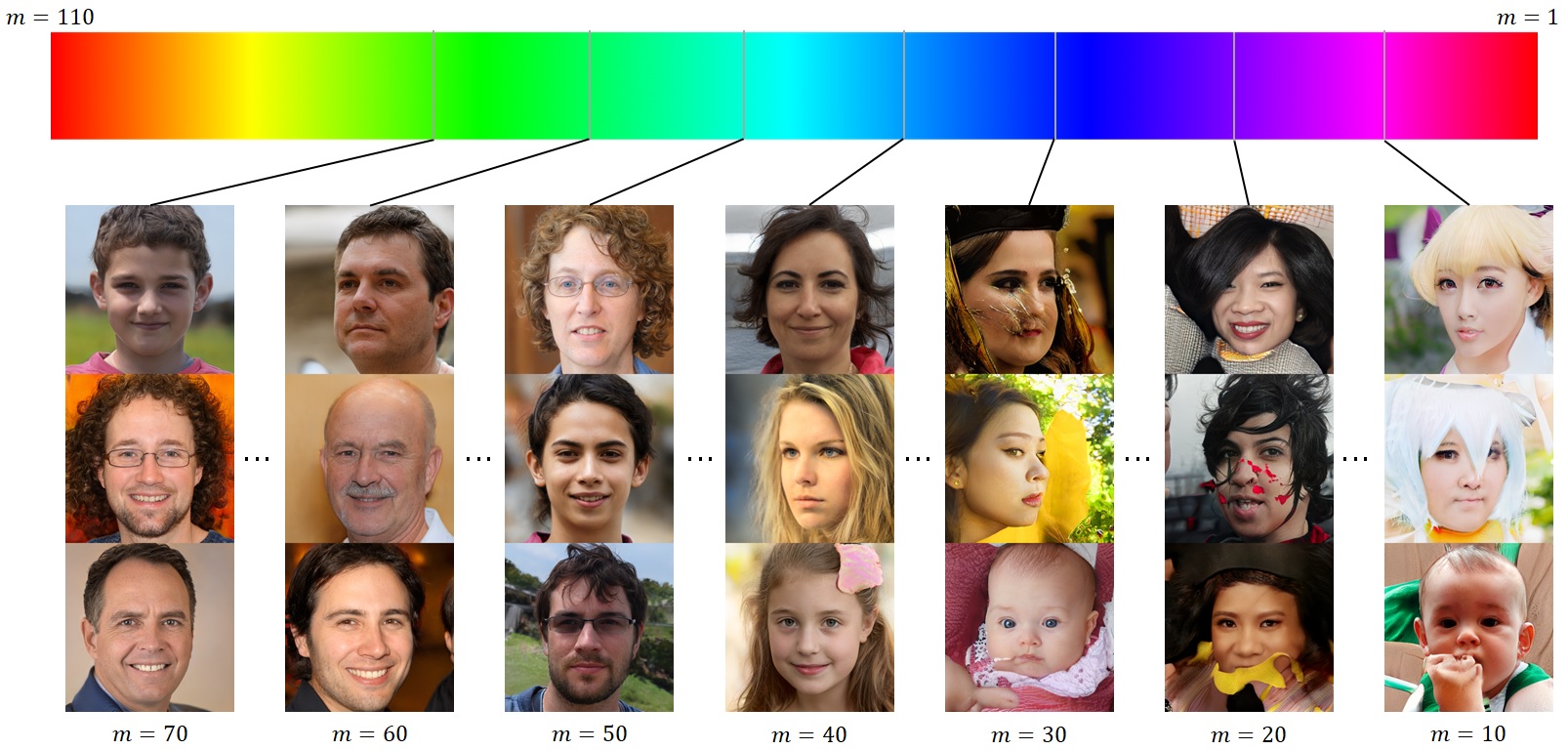}
    \caption{The {\it hubness spectrum} of StyleGAN2~\cite{karras2020analyzing} synthesized images ranked according to their {\it hub} values $m$. We use $n=10000$, $k=5$. 
    Note that the spectrum is highly tailed to the left and thus there are few images in the range $m=(70,110)$.
    }
    \label{fig:spectrum}
\end{figure*}

Fig.~\ref{fig:spectrum} shows the {\it hubness spectrum} obtained by our hubness priors. It can be observed that the quality of images changes from high to low from left to right with decreasing $m$.

\section{Example Images with Truncation Trick}
\label{appendix:truncation_trick_example}

\begin{figure}[htp]
\centering
    \centering
{\includegraphics[width=.9\linewidth]{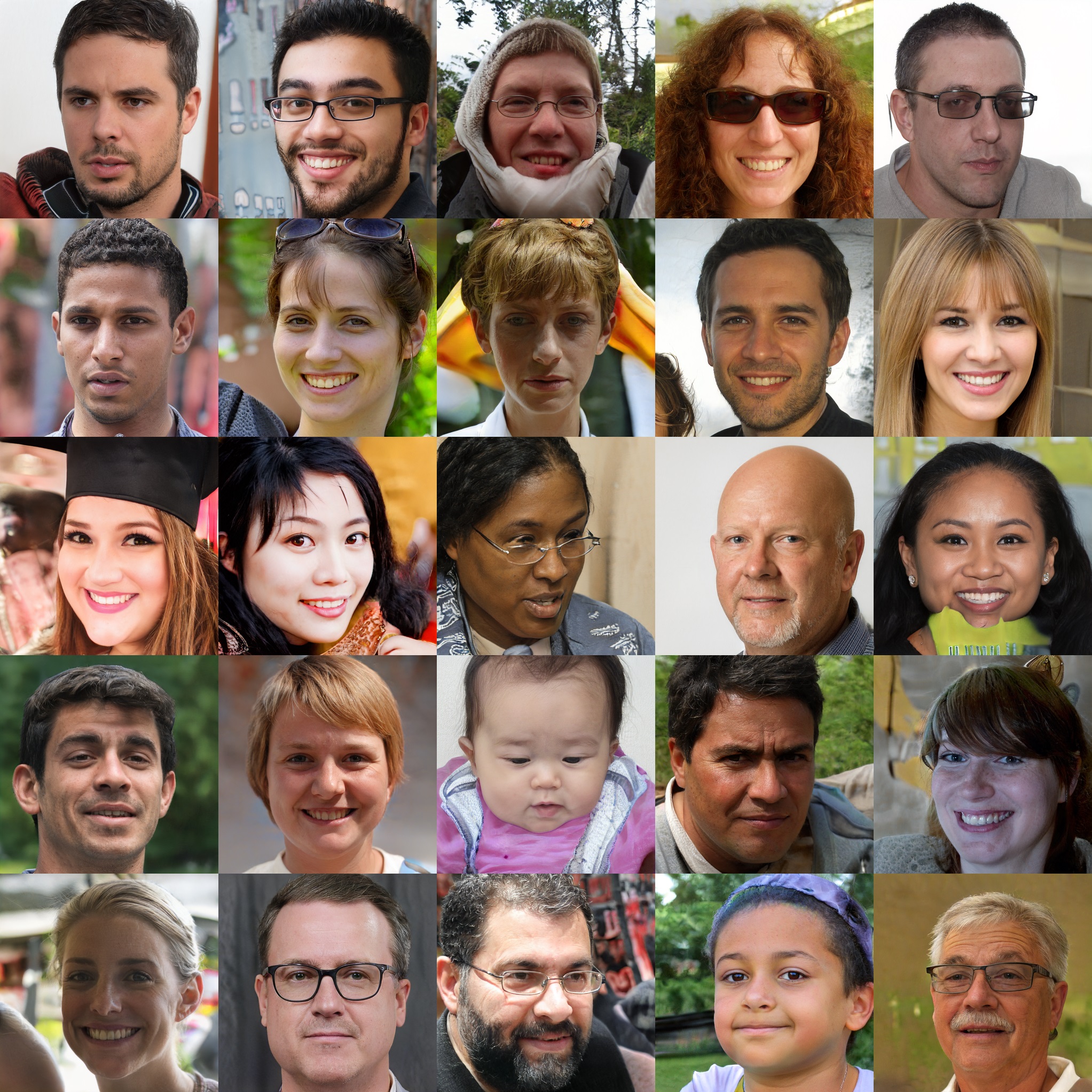}}%
    \caption{Examples of StyleGAN2 synthesized images after the truncation trick ($\psi = 0.7$).}%
    \label{fig:trunc_examp}%
\end{figure}

We show examples of StyleGAN2 synthesized images after the truncation trick ($\psi = 0.7$) in Fig.~\ref{fig:trunc_examp}.

\section{Running Time}
\label{appendix:running_time}

Table~\ref{table:time_algChoice} shows the running time of our method with different choices of $k$ and $n$. It can be observed that the running time increases mildly with $k$ but significantly with $n$.

\begin{table}[]
\centering
\vspace{-4mm}
\caption{Running time of our method using the StyleGAN2 $W$-space with different choices of $k$ and $n$. The default parameter values are $k=5$, $t=50$ and $n=10000$.}
\begin{tabular}{rr|rr}
\toprule
$k$  & Time(s) &   $n$      & Time(s)  \\ \midrule
$3$  & 163s &   $10000$  & 167s\\
$5$  & 167s &   $20000$  & 647s\\
$7$  & 176s &   $30000$  & 1272s\\
$10$ & 185s &   $40000$  & 2554s\\ \bottomrule
\end{tabular}
\vspace{-4mm}
\label{table:time_algChoice}
\end{table}

\section{Limitation and Future Work}

\begin{figure}[htp]
\centering
    \centering
    \includegraphics[width=.99\linewidth]{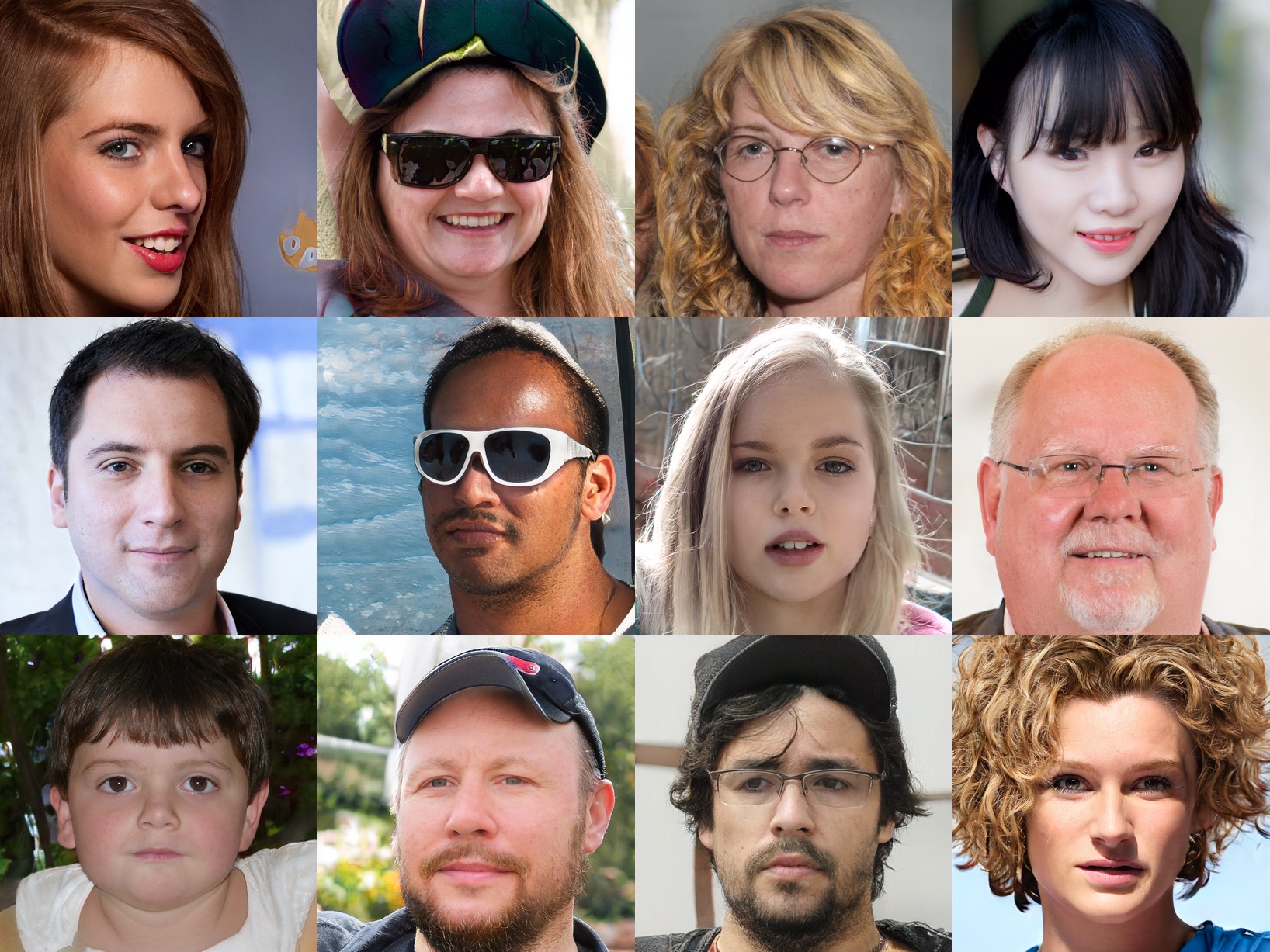}%
    \caption{Relatively high-quality StyleGAN2~\cite{karras2020analyzing} synthesized images with small hub values $m$. However, there are still small artifacts in these images ({\it e.g.} background and facial details).}
    \label{fig:boundaryResult_good}%
\end{figure}

Although our method allows for the sampling of high-quality latents, the quality of synthesized images is bounded by the performance of the pre-trained GANs used to synthesize them.
Also, we observed that the proposed {\it hubness priors} may overlook some relatively high-quality images with small {\it hub} values $m$ (Fig.~\ref{fig:boundaryResult_good}).
We conjecture that the reason might be that the limited sizes of latent sample sets ({\it e.g. $n=10000,20000,...$}) cannot capture all {\it hub} latents. This is partially verified by our experiment on the choice of $n$. However, it is difficult to test larger $n$ due to the $O(n^2)$ time complexity to compute the hub values $m$ for all points in a latent sample set. 
We hope to investigate this issue in future work.
We also hope to apply our insights on the hubness phenomenon in GAN latent space to improve the training of GANs and make GANs unbiased for all latents.
The acceleration of our algorithm is also a very interesting direction for future work.

\end{document}